\documentclass{article}

\usepackage[eandd, final]{neurips_2026}

\usepackage[utf8]{inputenc} % allow utf-8 input
\usepackage[T1]{fontenc}    % use 8-bit T1 fonts
\usepackage{hyperref}       % hyperlinks
\usepackage{url}            % simple URL typesetting
\usepackage{booktabs}       % professional-quality tables
\usepackage{amsfonts}       % blackboard math symbols
\usepackage{amsmath}
\usepackage{nicefrac}       % compact symbols for 1/2, etc.
\usepackage{microtype}      % microtypography
\usepackage[dvipsnames]{xcolor}         % colors
\usepackage[pdftex]{graphicx}
\setcitestyle{numbers,square}

\title{CME–SpectrumBench: Can LLMs Analyze Condensed Matter Spectral Data?}

\author{%
  Jin Gene Wong \\
  Physics Department and \\
  Stanford Laboratory for Intelligent Matter \\
  Stanford University \\
  Stanford, CA 94305 \\
  \texttt{jgw33@stanford.edu} \\
  \And
  Anjney Midha \\
  Applied Physics Department \\
  Stanford University \\
  Stanford, CA 94305 \\
  \texttt{anjney@stanford.edu} \\
  \And
  Joseph Mathew Tennyson \\
  Electrical Engineering \\ \& Computer Sciences \\
  University of California, Berkeley \\
  Berkeley, CA 94720 \\
  \texttt{josephtennyson@berkeley.edu} \\
  \And
  Wei-Lin Chiang \\
  Electrical Engineering \\ \& Computer Sciences \\
  University of California, Berkeley \\
  Berkeley, CA 94720 \\
  \texttt{weichiang@berkeley.edu} \\
  \And
  Ion Stoica \\
  Electrical Engineering \\ \& Computer Sciences \\
  University of California, Berkeley \\
  Berkeley, CA 94720 \\
  \texttt{istoica@berkeley.edu} \\
  \And
  Zhi–Xun Shen \\
  Applied Physics Department, \\
  Physics Department, and \\
  Stanford Laboratory for Intelligent Matter \\
  Stanford University \\
  Stanford, CA 94305 \\
  \texttt{zxshen@stanford.edu} \\
}

\begin{document}

\maketitle

\begin{abstract}
  Large Language Models (LLMs) are presently well assessed by expert–level mathematics and physics benchmarks which focus on \textit{theoretical}, symbolic reasoning. However, we lack \textit{experimental} benchmarks that evaluate data analysis in frontier research problems, a capability that future domain–specific foundation models must have. To this end, we present \textbf{CME–SpectrumBench}, an expert–designed benchmark consisting of 190 questions that measures how reliably frontier LLMs analyze research–level spectral scientific data in condensed matter experiments (CME). Using simulated CME spectral data, we prompt models to identify structures from 2D intensity arrays in the face of noise and limited instrumental resolution, a test of dense context reasoning. Models are scored by how accurately their numerical answers approach the ground truth. Under direct query, scores generally pale in comparison to handwritten code due to clear failure modes: models can find simple features but struggle to trace patterns or compute derived quantities, and performance typically degrades with increased data resolution and noise/convolution levels. LLM performance becomes comparable to handwritten code by allowing tool use (access to a Python interpreter) with/without ReAct prompting, in effect mapping questions onto agentic code generation problems. Our findings show that frontier models can serve as research assistants \textit{by interfacing with data through code}, but still face difficulties \textit{processing information directly from large datasets} — a key barrier future foundation models must overcome. The code and data are publicly available at \href{https://huggingface.co/datasets/jingene/CMESpectrumBench}{\textcolor{blue}{\texttt{https://huggingface.co/datasets/jingene/CMESpectrumBench}}}.
\end{abstract}

\section{Introduction} \label{section1}

As frontier AI systems are increasingly adopted in the sciences, we need evaluations for rigorous, dependable scientific reasoning. Many scientific benchmarks focus on competence in code generation \citep{chen2024}, workflow generation \citep{majumder2024}, or symbolic reasoning \citep{pan2025, wang2025a, wang2025}, drawing on insights by earlier, non–scientific benchmarks \hyperref[appendixA]{(Appendix A)}. However, a core skill for scientific reasoning remains relatively unevaluated: extracting meaningful information from noisy experimental data, a task which scientists, AI agents, and domain–specific foundation models must perform reliably.

\textbf{CME–SpectrumBench} fills this gap. It is the first benchmark to test frontier models on research–level spectral analysis in condensed matter \textit{experiments} (CME), a necessary counterpart to \textit{theoretical} condensed matter benchmarks developed in the past year \citep{pan2025, wang2025a, wang2025}. Condensed matter, the field responsible for developing advanced materials underpinning modern technology—including computing hardware enabling the advent of AI itself—promises to beget further scientific and computing breakthroughs \hyperref[appendixB]{(Appendix B)}. It shares, with AI, the common thrust of studying \textit{extremely complex systems} (on the order of Avogadro's constant, $\sim10^{23}$, see \hyperref[appendixD]{Appendix D}), but lacks the breadth of features afforded to LLMs, \textit{viz.} interpretability and Golden Gate Claude \citep{templeton2024}, \textit{because these features are accessible only if an experiment is physically feasible}. The most direct, information–dense experimental tools are \textbf{spectroscopic probes}, the tip of the spear in condensed matter research. CME–SpectrumBench focuses on Angle–Resolved Photoemission Spectroscopy (ARPES) \citep{damascelli2003, sobota2021} and Resonant Inelastic X–ray Scattering (RIXS) \citep{ament2011, mitrano2024}, two well–characterized spectroscopies.

\begin{figure}[h!]
  \centering
  \includegraphics[width=0.9\linewidth]{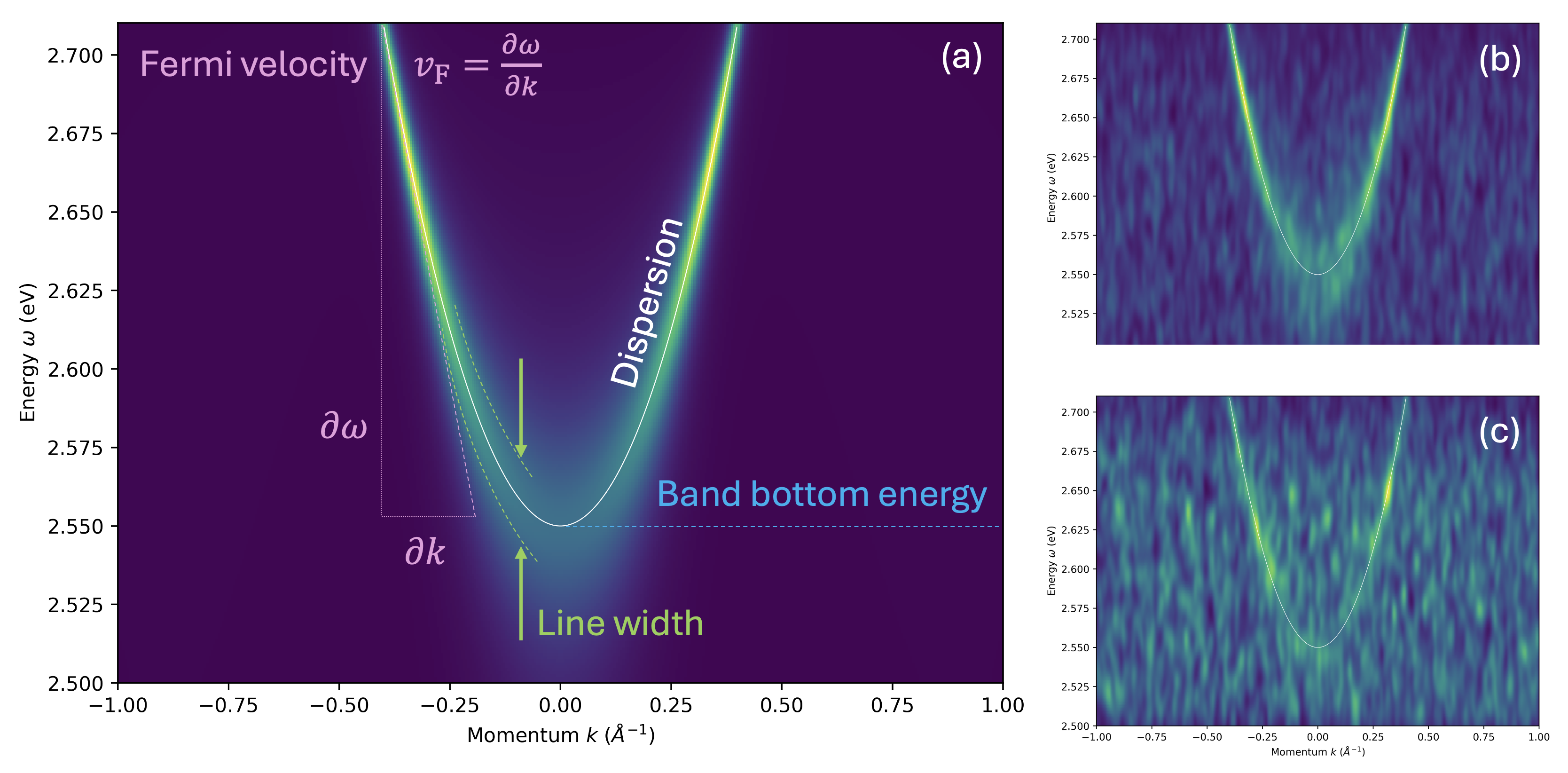}
  \caption{Sample spectral function $A(k,\omega)$ from the bottom of a band, from which we may draw three tiers of questions \hyperref[section2.3]{(Section 2.3)}. Increasing noise levels are shown: \textbf{(a)} $0\%$, \textbf{(b)} $10\%$, \textbf{(c)} $40\%$.}
  \label{fig:example}
\end{figure}

Spectral analysis is a \textbf{long–context inverse problem} that shares key attributes with dense context reasoning (MMReD) \citep{kurkin2025}, Multi–Round Co–reference Resolution (MRCR) \citep{openai2025}, and latent structure queries (Michelangelo) \citep{vodrahalli2024}. Transcending prior \textbf{`needle in a haystack'} (NIAH) tasks where models \textit{retrieve limited information sparsely embedded in a long and noisy context}, in these evaluations models must \textit{integrate \textbf{extensive information} densely dispersed across an entire context} (`$x$') in order to \textit{retrieve \textbf{abstract structures}} (`$y$'). Likewise, information needed to calculate analytic quantities in CME spectra (`$y$') are diffused across large amounts of data (`$x$') complicated by experimental artefacts. For example, in Fig.~\ref{fig:example}(a), the spectrum (`$x$') contains information (`$y$') extracted by analyzing systematic patterns—such as its dispersion (solid white line) or line width (which broadens the dispersion)—despite corrupting effects of noise and convolution due to instrumental resolution, shown in Fig.~\ref{fig:example}(b,~c). The ability to perform dense context reasoning is a prerequisite for future foundation models in physics. Whereas biological foundation models \citep{ji2021, zhou2024, nguyen2024, brixi2026, hayes2025} have \textit{discrete} tokenizable elements (such as amino acids or base pairs, akin to scripts in natural languages), data in physics are \textit{continuous}, albeit binned into 1D intervals, 2D pixels, or 3D voxels. This attribute introduces an additional complication for foundation models in physics: \textit{it is unclear how data should be tokenized to balance computational efficiency and quantitative accuracy}, a question which stands in addition to the \textit{extant} problem of models learning patterns from priorly tokenized elements (such as biological models learning protein structures from amino acid sequences).

\begin{figure*}[h!]
  \centering
  \includegraphics[width=\linewidth]{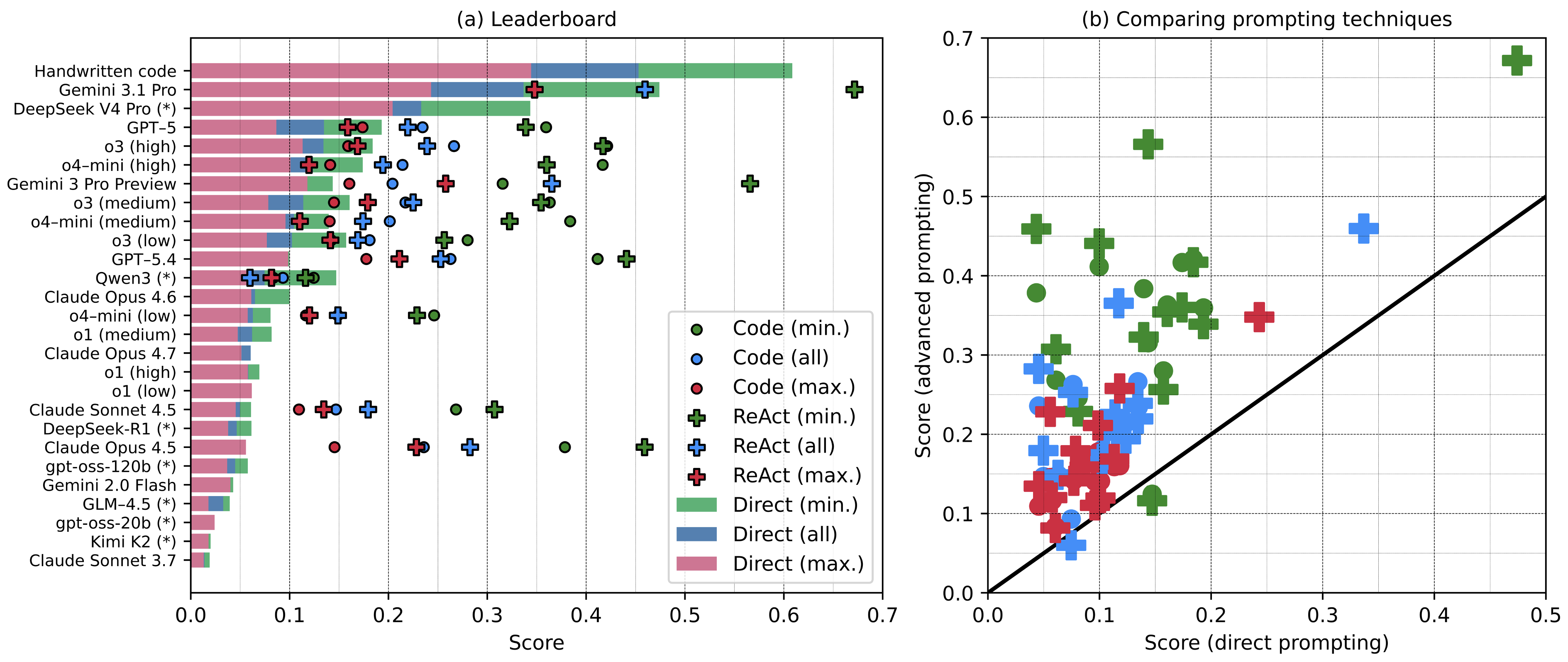}
  \caption{\textbf{(a)} More recently released, closed–weight (proprietary) models generally perform better. Models with asterisks (*) are open–weight. Note expanded axes scaling, which top out at 1. \textcolor{ForestGreen}{Min.}/\textcolor{BrickRed}{max.} refer to spectra with minimum and maximum noise or convolution. \textbf{(b)} Advanced prompting improved scores, particularly for questions with minimal noise and convolution \textcolor{ForestGreen}{(min.)}, but gains were marginal when experimental complications were severe \textcolor{BrickRed}{(max.)}. A solid black line denotes identical scores under direct query and advanced prompting. The legend applies to both subfigures.}
  \label{fig:leaderboards}
\end{figure*}

In CME–SpectrumBench, models extract patterns by learning from examples rather than applying explicitly stated rules, and numerical answers are scored by their deviation from the ground truth. High–resolution datasets pushing context length limits were generated by a high–fidelity ARPES and RIXS spectrum simulator to avoid training contamination. \textit{We test recent reasoning models and find that only frontier systems released after December 2024 show meaningful progress, as shown in the leaderboard in} \textbf{Fig.~\ref{fig:leaderboards}}. (As a rule of thumb, a score of 1 is a perfect answer, while 0.5 deviates by an amount equal to typical experimental standard deviation, $1\sigma$.) Although newer models perform better, when vanilla \textbf{direct prompting} is used, the best model achieves 0.34 out of 1 \textcolor{NavyBlue}{averaged over all spectra (blue bars)}, rising to 0.48 for questions with \textcolor{ForestGreen}{minimal noise/convolution (green bars)}. Enabling \textbf{code generation} with/without \textbf{ReAct prompting} increased scores to as much as 0.46 \textcolor{NavyBlue}{(all spectra)} and 0.68 \textcolor{ForestGreen}{(minimum noise/convolution)}, with strong gains for questions with relatively low noise/convolution, but minimal improvement when severe (Fig.~\ref{fig:leaderboards}b). Within the latest slew of open–weight models, DeepSeek V4 Pro was by far the best, scoring 0.23 \textcolor{NavyBlue}{(averaged)} and 0.35 \textcolor{ForestGreen}{(minimum noise/convolution)}. In comparison, several hundred lines of basic code written by a graduate student over three days, with input from researchers of equivalent calibre (\hyperref[appendixF]{Appendix F} and repository), achieve 0.45 \textcolor{NavyBlue}{(all spectra)} and 0.61 \textcolor{ForestGreen}{(minimum noise/convolution)}, a significant gap in performance over all but the very best of models.

\textbf{We highlight three key contributions of our benchmark:}\begin{enumerate}
    \item CME–SpectrumBench is the first to test rigorous \textbf{experimental data analysis} by LLMs, under closed–book, three–shot conditions, in the scientifically and technologically critical field of \textbf{condensed matter}. Questions may be made as difficult as desired by reducing tolerance to deviation, or by extending token counts via higher resolution data.
    \item Our expert–designed synthetic dataset encompasses \textbf{actual research problems} with \textbf{realistic experimental complications}. These problems unveiled common failure modes: models can find simple features but struggle to trace patterns or compute derived quantities, and performance degrades with increased data resolution and noise/convolution levels.
    \item We demonstrate competitive model performance by converting questions into \textbf{agentic code generation problems} (enabling code interpreter access/ReAct prompting), thus frontier models can serve as research assistants for data analysis. However, \textbf{poor performance remains on direct query dealing with long context}, which must be overcome by future \textit{foundation models handling noisy, massive datasets natively in scientific domain languages}.
\end{enumerate}

\section{CME–SpectrumBench: Design \& Construction} \label{section2}

\subsection{Attributes of ARPES \& RIXS spectral data} \label{section2.1}

CME–SpectrumBench tests how models extract quantitative information from scientific datasets with realistic complications: noise and instrumental convolution. We focus on Angle–Resolved Photoemission Spectroscopy (ARPES), which measures electronic structure in materials, and Resonant Inelastic X–ray Scattering (RIXS), which measures elementary excitations such as phonons, plasmons, and magnons. Their spectral data are the `domain languages' of condensed matter \textit{least removed from the theoretical ground truth, with the richest feature space}—their data are as fundamental to materials as the genomic code is to biology—\textit{thus these are among the most important experiments for condensed matter physics} (further reading in \hyperref[appendixB]{Appendix B}).

Each dataset is a 2D intensity map with energy ($\omega$) and/or momentum ($k$) axes. (Note that $\omega$– and $k$–space representation is the natural language of the field, due to the wave–like nature of electrons and photons under quantum mechanics.) Figure~\ref{fig:example}(a) shows an example. Pixels represent electron count rates across energy and momenta (the spectrum), which reveals electronic behavior in the material: brighter regions show where electrons are more likely to be found. Physicists extract dispersion curves $\epsilon(k)\in\mathbb{R}$ (how electron energy varies with momentum) and line widths $\Sigma''(\omega)\in\mathbb{R}$ (how broad spectral features are at different energies). Spectral analysis is then an \textbf{inverse problem}. We know the \textit{forward process}: given $\epsilon(k)$ and $\Sigma''(\omega)$, we can compute the spectrum,
\[A(k,w) = \frac{\Sigma''}{(\omega - \epsilon)^2 + (\Sigma'')^2}.\]
In the language of \hyperref[section1]{Section 1}, this equation maps $y \mapsto x$, where the spectrum $A(k,\omega)$ is the quantity measured (`$x$'), while dispersion $\epsilon$ and line width $\Sigma''$ are ground–truth quantities we wish to extract (`$y$'). Data analysis aims to work out the \textit{inverse problem} $x \mapsto y$, which is harder (`given a noisy spectrum, extract the underlying dispersion and line width functions').

ARPES and RIXS spectra are produced by a \textbf{spectrum simulator} written expressly for this benchmark, included in the dataset. \textit{Instructions to run the simulator are stated in \hyperref[appendixF]{Appendix F} under \hyperref[rep]{Reproducibility}.} We note several strengths of evaluating on ARPES and RIXS: (1) a clear, built–in ground truth exists, (2) experimental noise can be controlled, (3) difficulty is adjustable, and (4) `correctness' is not binary but a continuous score parameterized by deviation from the ground truth.

\subsection{Structure of dataset and questions} \label{section2.2}

38 question categories \hyperref[appendixF]{(Appendix F)} were generated by the spectrum simulator, each prepared at five noise or convolution levels, for a total of 190 questions. We use few–shot prompting (almost all three–shot): models read example spectra (with answers provided) to figure out an analysis method, then analyze a spectrum for the test itself. Under \textbf{direct prompting}, questions were stated as text strings, beginning with the prompt and followed by example and test spectral datasets. All prompts contain the minimal amount of information and concise definitions for models to comprehend each problem, and caution about the presence of experimental complications (such as noise and convolution), while refraining from prescribing particular analysis methods. 

Example and test datasets are contained in a single text string for each question. The first row of each dataset (after `\texttt{Energy\,(eV)/Momentum:}') states the momentum corresponding to each column; the first (leftmost) column states the energy of each row. Remaining entries list spectral intensities per pixel for the range of momenta and energies contained. To reduce token count while retaining a high dynamic range, spectral intensities are normalized such that the highest intensity is exactly 1000, and all intensities are rounded to the nearest integer. An example is shown:

\textcolor{gray}{\hrulefill}

\textcolor{gray}{Four datasets showing ARPES spectra are contained. They are labeled ``Dataset A'', ``Dataset B'', ``Dataset C'', and ``Dataset D''. Read ``Dataset A''. The Fermi energy of ``Dataset A'' is 2.71 eV. Read ``Dataset B''. The Fermi energy of ``Dataset B'' is 15.98 eV. Read ``Dataset C''. The Fermi energy of ``Dataset C'' is 8.01 eV. Now read ``Dataset D''. State the Fermi energy of ``Dataset D'' in units of electron–Volts. Print only your numerical answer.}

\textcolor{gray}{Dataset A}

\textcolor{gray}{Energy (eV) / Momentum:\quad $-1$\quad	$-0.973$\quad	$-0.9459$\quad	$-0.9189$\quad	$-0.8919$\quad	\ldots}

\textcolor{gray}{2.5\quad	513\quad	622\quad	561\quad	609\quad	633\quad	753\quad	619\quad 685\quad	513\quad	566\quad	727\quad	520\quad	493\quad	548\quad \ldots}

\textcolor{gray}{2.504\quad	566\quad	644\quad	600\quad	654\quad	671\quad	818\quad 667\quad	760\quad	544\quad	557\quad	811\quad	560\quad	514\quad \ldots}

\textcolor{gray}{\ldots}

\textcolor{gray}{\hrulefill}

Advanced prompting was also used on better performing models. Separate tests were run where we have enabled access to a \textbf{code interpreter}, where datasets were also provided in the form of \textbf{tabular .csv files} rather than as text strings in the query itself. An additional set of tests were performed, where models \textit{also} used \textbf{ReAct prompting} \citep{yao2023} as a code debugging mechanism, by being prompted to compare example answers with those obtained from running example datasets through their code. Although reasoning traces were recorded, only final numerical answers were scored.

\subsection{Types of question} \label{section2.3}

The 38 categories of questions \hyperref[section2.2]{(Section 2.2)} are grouped under seven scientific domains. Because this classification is primarily relevant to physicists, we leave fuller explanations in \hyperref[appendixF]{Appendix F} and the captions of Figs.~\ref{fig:a}--\ref{fig:g}. In parallel, we classify questions into three broad \textbf{difficulty tiers}: \begin{itemize}
    \item \textbf{Tier I.} \textit{Extracting} a \textbf{scalar} quantity, departing least from NIAH questions.
    \item \textbf{Tier II.} \textit{Extracting} an \textbf{array} of quantities, such as dispersions $\epsilon(k)$ and line widths $\Sigma''(\omega)$.
    \item \textbf{Tier III.} \textit{Indirectly determining} \textbf{scalar} quantities (calculated after extracting Tier II arrays).
\end{itemize} \textbf{The spectrum shown in Fig.~\ref{fig:example}(a) is illustrative.} It may be used to ask questions from three tiers of difficulty. Asking for the scalar band bottom energy (blue line at bottom of parabola) is the easiest \textbf{(Tier I)}. Tracing the dispersion (white line) or line width (green lines) as arrays are \textbf{Tier II}. Asking for the scalar Fermi velocity $v_{F}$, the gradient $\partial \omega/\partial k$ at the top, \textit{is in principle the most complicated} \textbf{(Tier III)}, answered \textit{rigorously} only by first tracing the dispersion \textbf{(Tier II)} to then calculate $v_{F}$.

\subsection{Real--world complications: noise and convolution} \label{section2.4}

\textbf{Noise} was inserted in the manner of Fig. 6(a) in \citet{kim2021a}: as a set of $2\times10^4$ (ARPES) or $5\times10^3$ (RIXS) randomly–distributed spots, their intensities randomly chosen within a range, and footprints broadened into 2D Gaussians. The amount of noise is defined as \textit{mean} noise intensity as a fraction of \textit{maximum} spectral intensity (prior to adding noise). We do not measure against mean spectral intensity, as it depends on the size of the feature investigated relative to the size of the dataset, while maximum spectral intensity does not. We are interested in extracting strong signals regardless of how much other information (the background signal) is present. Figs.~\ref{fig:example}(a–c) show representative noise levels; for most question categories, noise levels scale up to $40\%$.

Additionally, all data were \textbf{convolved} with 1D Gaussians of width $0.005/\textup{\AA}$ in momentum, and $0.003$ eV in energy. This is present in all real--world experimental data due to instrumental convolution. At such a low level of convolution, only minor artefacts such as slight deviations of observed dispersions from the ground truth are produced. Variable convolution levels are also posed in RIXS questions.

\textbf{The presence of noise and convolution do not alter the ground truth.} For example, the ground truth of array–type questions (Tier II) are `smooth' regardless of noise and convolution. The `smooth' ground truth of example spectra are given to models to guide their reasoning; we expect optimal answers to be `smooth' regardless of how much noise/convolution is present. Again, we reinforce the link with \textbf{dense context reasoning}, such as in MMReD \citep{kurkin2025}, which requires models to reason across extended contexts whose full contents must be processed rather than ignored.

\section{Evaluation Protocol \& Metric} \label{section3}

\subsection{Few–shot prompting} \label{section3.1}

Dependence on \textbf{elicitation} is reduced by few–shot prompting, so that performance would not depend on the skill of the evaluator rather than the model. Although models place higher importance on information at the start of prompts \citep{liu2024}, the vast majority of content by token count are the examples themselves ($\sim10^5$ tokens) rather than worded starting instructions (at most a few hundred tokens). Additionally, in preliminary tests, we had observed that models sometimes defaulted to \textbf{incorrect heuristics}: they regurgitated answers of examples rather than reasoned to obtain the answer of the actual question. We have therefore ensured that answers did not overlap with those of prior test examples, so that regurgitation does not accidentally inflate the score. Lastly, we note that dependence on the \textbf{number of examples} could not be investigated independently of token count/resolution: because of large token count, varying the number of examples would introduce trade–offs, either reducing resolution, or performance scaling with token count (see \hyperref[effectofres]{Effect of resolution}, Section 4.3).

\subsection{Evaluation metric: Lorentzian scoring function} \label{section3.2}

Responses are not scored binarily (strictly correct or incorrect). Instead, we score each response using a rescaled \textbf{Lorentzian}, parameterizing their numerical deviation from the ground truth. For a single scalar answer, this takes the form (treated in \hyperref[appendixE]{Appendix E}) \[\textrm{Score} = \frac{\gamma^2}{(x - x_0)^2 + \gamma^2}\]for model response $x$, ground truth $x_0$, and half--width half--maximum (HWHM) $\gamma$. A completely correct response is scored 1, and a completely \textit{\textbf{in}}correct response 0. As a rule of thumb, an answer deviating by typical experimental standard deviation (HWHM $\gamma$) scores 0.5. A clear ground truth exists for each question, whether as a single number or an array of numbers — outputs of an analytic function across an energy ($\omega$) or momentum ($k$) domain. \textit{For array responses, each element is independently scored in the same manner, then their scores were averaged.} HWHM $\gamma$ takes the values of $0.005/\textup{\AA}$ in momentum, $0.003$ eV in energy, and $0.01$ in doping (see Fig.~\ref{fig:e} of \hyperref[appendixG]{Appendix G}) comparable to standard deviation ($1\sigma$) in experimental data analysis today. Note that the test may be made arbitrarily difficult by decreasing $\gamma$ (error tolerance) to reduce all scores monotonically.

Models are prompted to return a number or set of numbers. No numerical restriction is imposed on their response other than the expected array length, which is clearly stated at the start of each question (answers with incorrect lengths are discarded). Each model's score is first averaged amongst multiple responses to the same question, then averaged over all 190 questions.

\subsection{Models evaluated} \label{section3.3}

Preliminary tests showed that non–reasoning models (prior to December 2024) performed poorly on our test. Thus, we restricted our evaluation to reasoning models released December 2024 and after. Firstly, this includes various \textbf{closed–weight models} from OpenAI (o1, o3, o4–mini, GPT–5, and GPT–5.4), Google (Gemini 2.0 Flash, Gemini 3 Pro Preview, Gemini 3.1 Pro), and Anthropic (Claude Sonnet 3.7 and 4.5; Opus 4.5, 4.6, and 4.7). OpenAI models with adjustable inference–time compute were evaluated as a function of compute mode (low, medium, and high). Secondly, \textbf{open–weight models} were tested: DeepSeek–R1, Qwen3 (on Thinking Mode), Kimi K2, GLM-4.5, gpt–oss–120b, gpt–oss–20b, and DeepSeek V4 Pro. Leaderboard results are shown in Fig.~\ref{fig:leaderboards}. Exact model names and their context window limits are stated in \hyperref[rep]{Reproducibility} in \hyperref[appendixF]{Appendix F}.

\section{Results: agentic advance and long context deterioration} \label{section4}

\textit{Responses generated by frontier models are included in \href{https://huggingface.co/datasets/jingene/CMESpectrumBench}{\textcolor{blue}{this repository}}, along with scripts required to reproduce Figs.~\ref{fig:leaderboards}, \ref{fig:tiers}, and \ref{fig:ablations} in particular. Instructions are given in the corresponding section on \hyperref[rep]{Reproducibility} in \hyperref[appendixF]{Appendix F}.}

\subsection{Comparing models} \label{section4.1}

For fairness, we first tested models with the \textbf{same standard resolution} of $100^2$ data points, the highest resolution accepted by models with the smallest (low resolution) context windows. The correlation between release date and model performance is clear (Fig.~\ref{fig:leaderboards}), demonstrating progress in the performance of reasoning models in 2025/26. Under \textbf{direct query}, closed–weight models by OpenAI, Google, and Anthropic are among the best performing; these likely reflect the long inference times and long built--in context windows that the models afford. Among open–weight models, DeepSeek V4 Pro performed competitively. We note a gradual advance in performance in models from December 2024 to February 2026, and a substantial leap by two models released thereafter (Gemini 3.1 Pro and DeepSeek V4 Pro). Full scores for direct query are shown in Fig.~\ref{fig:direct}.

\begin{figure}[h!]
  \centering
  \includegraphics[width=\linewidth]{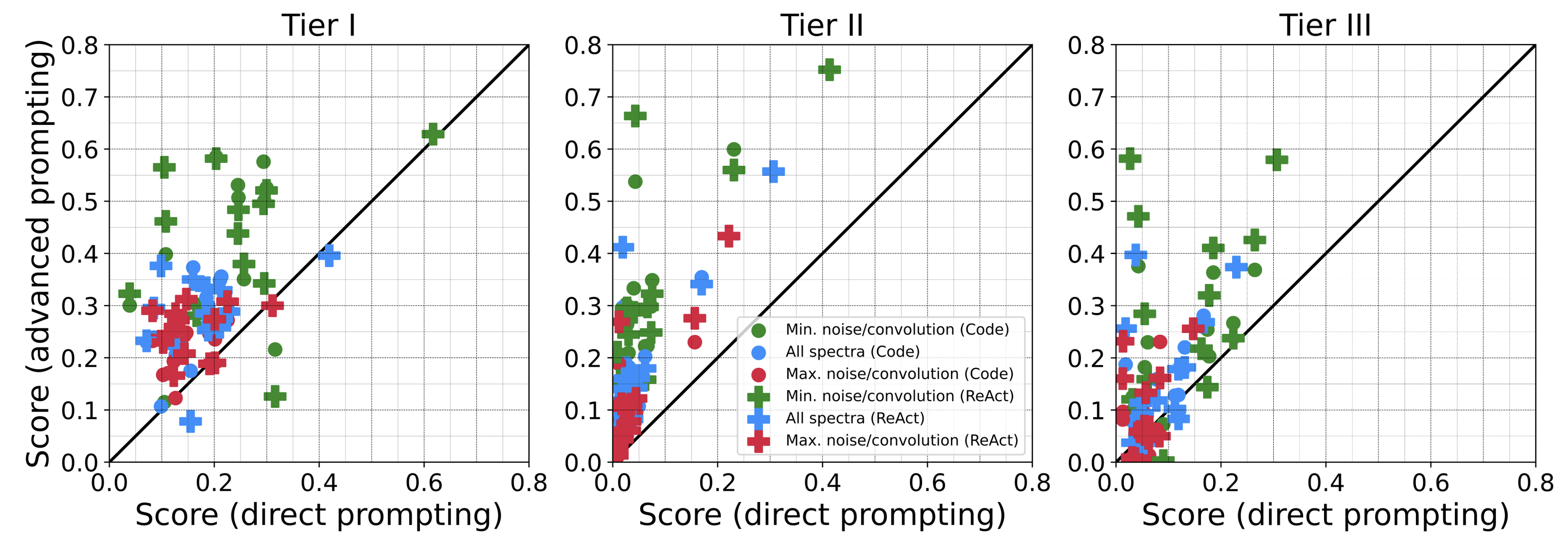}
  \caption{Advanced prompting improved scores slightly for Tier I and Tier III questions, and vastly for Tier II. However, questions with maximal noise or convolution \textcolor{BrickRed}{(red)} were still poorly tackled.}
  \label{fig:tiers}
\end{figure}

Access to a \textbf{Python interpreter} was additionally given to several well–performing models. Tabular data was presented in .csv and models prompted to produce code subsequently input to the interpreter, whose numerical output were scored. As a test of their agentic capabilities, \textbf{ReAct prompting} \citep{yao2023} was separately used (Fig.~\ref{fig:leaderboards}(b) and Fig.~\ref{fig:advanced}). Gains in score were substantial at low noise or convolution, but marginal when these experimental complications were severe (Fig.~\ref{fig:tiers}). Some models (Claude Opus and Sonnet 4.5) performed much better under code/ReAct than direct query, while little change was seen in others (Qwen3). Full scores for advanced prompting are shown in Fig.~\ref{fig:advanced}.

\subsection{Comparing tiers of questions} \label{comparingtiers}

Models performed best in \textbf{Tier I} questions, which are the simplest yet have already departed from NIAH measures by diffusing the `needle' (quantitative information) through the context (spectra). A soft `floor', or lower bound in score (around 0.1) is evident in this tier but not the others. Poorer performance was observed for \textbf{Tier II} (grasping an underlying latent structure) and in \textbf{Tier III} questions — whose \textit{proper, rigorous} analysis \textit{should} involve two steps models were not explicitly guided through: first retrieving a latent structure (Tier II), then calculating a quantity from the retrieved structure, such as Fermi velocity $v_F$ (which is the dispersion gradient, $\partial \omega/\partial k$, Fig.~\ref{fig:example}) or doping level ($k$– or momentum area enclosed at zero energy, see Fig.~\ref{fig:e}). Therefore, Tier III questions \textit{should in principle} be at least as difficult as Tier II. However, models oftentimes obtained higher Tier III scores than Tier II, suggesting that they have \textit{heuristically `short–circuited' the reasoning process}.

Under \textbf{advanced prompting}, that is, enabling access to a Python interpreter with/without ReAct prompting, scores almost uniformly increased. \textit{Scores for each model} are shown in Fig.~\ref{fig:tiers} \textit{for each tier}. Nevertheless, improvement was minimal when a severe amount of noise/convolution was present, indicating difficulty in mapping these questions into code generation problems, at least under the framework of our three–shot, closed–book evaluation. The strongest increase in scores were apparent for Tier II questions, suggesting that part of the difficulty in direct query (as opposed to advanced prompting) stemmed from retrieving long, array–like quantities (rather than scalars in Tiers I\,/\,III).

\begin{figure}[h!]
  \centering
  \includegraphics[width=\linewidth]{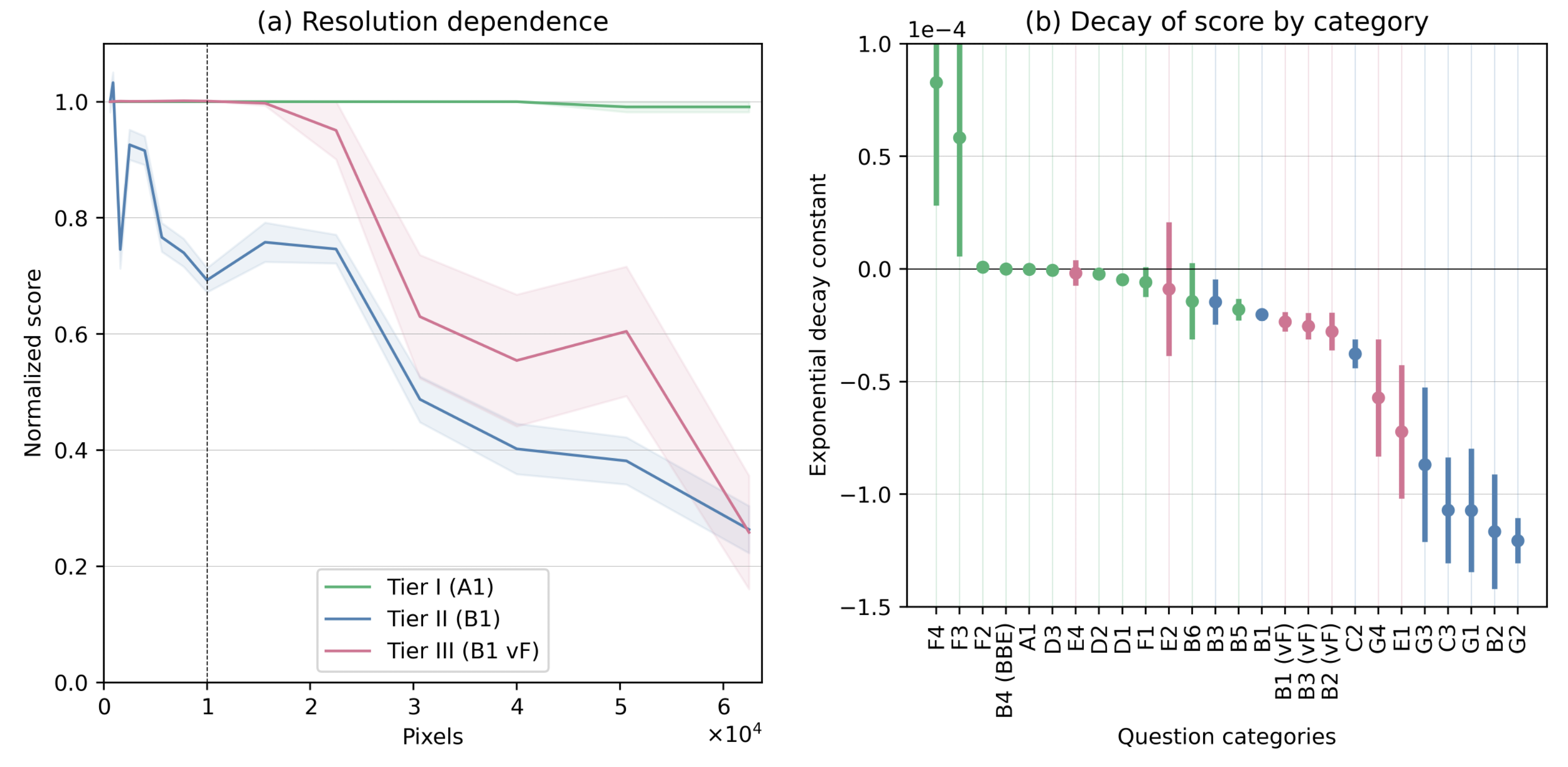}
  \caption{\textbf{(a)} Higher resolution (more pixels/token count) typically worsened scores. Three examples are shown, representing the three tiers of questions. A vertical dashed line at $100^2$ pixels indicates the resolution at which the standard evaluation is carried out in Fig.~\ref{fig:leaderboards}. Shaded areas indicate error in the mean score (standard error) $\sigma_{\overline{x}} = \sigma_x/\sqrt{n}$, for score $x$, sample standard deviation $\sigma_x$, and sample size $n$. \textbf{(b)} \textit{Maintaining the same color scheme}, exponential decay of score with resolution (beyond $100^2$ pixels) is shown across categories, highlighting the differences between tiers of questions.}
  \label{fig:ablations}
\end{figure}

\subsection{Effect of resolution} \label{effectofres}

\textbf{Decline in performance with context length} has long been observed in other long–context tasks \citep{kuratov2024, hsieh2024, lee2024} for which log–linear \cite{wang2025b} or sigmoidal \cite{zhou2025} decay trends were posited (sigmoidal decay asymptotes to log–linear decay in the long–context limit). To ascertain the extent to which this trend held in our evaluation, we carried out ablation tests on \textbf{dataset resolution} (that is, \textbf{token count}) in Gemini 3.1 Pro \textbf{under direct query}. Decline in score (\textit{normalized to the score at lowest resolution}) for three noiseless questions are shown in Fig.~\ref{fig:ablations}(a), exemplifying performance typical of three tiers of questions \href{section2.3}{(Section 2.3)} — \textcolor{ForestGreen}{\textbf{Tier I}} questions showed little or no decline with increasing resolution, \textcolor{NavyBlue}{\textbf{Tier II}} the steepest and earliest decline, and \textcolor{BrickRed}{\textbf{Tier III}} an intermediate rate of decrease. Fairer–than–expected performance of \textit{principally} `harder' \textcolor{BrickRed}{\textbf{Tier III}} compared to \textcolor{NavyBlue}{\textbf{Tier II}} at high resolution (long context) may be ascribed either to heuristic shortcuts mentioned in \hyperref[comparingtiers]{Section 4.2}, or the relative ease of retrieving scalar quantities in \textcolor{BrickRed}{\textbf{Tier III}} as opposed to arrays in \textcolor{NavyBlue}{\textbf{Tier II}}. Gathering fits to \textbf{exponential decay} of score in the long context limit ($100^2$ pixels and beyond), we show that these observations hold for other categories—see Fig.~\ref{fig:ablations}(b) on noiseless questions—spurious increases in score for two \textcolor{ForestGreen}{\textbf{Tier I}} questions (\texttt{F3} and \texttt{F4}) notwithstanding.

\subsection{Effect of noise} \label{Section4.5}

\begin{figure}[h!]
  \centering
  \includegraphics[width=\linewidth]{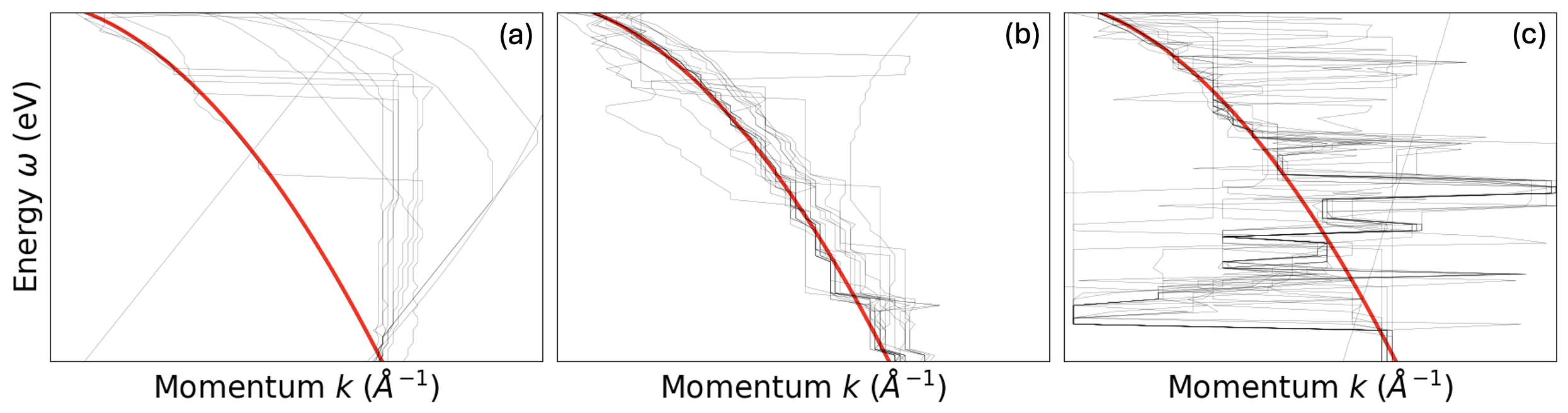}
  \caption{Unexpected improvement and subsequent expected decline in accuracy of answers with increasing noise; ground truth in red. Noise levels are set at \textbf{(a)} $0\%$, \textbf{(b)} $10\%$, and \textbf{(c)} $40\%$ of maximum dispersion intensity, for a quadratic dispersion (question category \texttt{B2}).}
  \label{fig:b2}
\end{figure}

Introducing \textbf{intensity noise} is analogous to increased perceptual complexity, as termed by MMReD \citep{kurkin2025}, where synthetic occlusion–like errors render long context problems harder to solve. Although scores clearly depreciate with increasing noise (Fig.~\ref{fig:leaderboards}), no obvious, universal functional form may be fit to the trend. Nevertheless, some qualitative observations emerge, as shown in Fig.~\ref{fig:b2}, which tested Gemini 2.5 Pro Preview on question \texttt{B2} at varying noise levels. In general, increasing noise visually produced an `injection of randomness' in responses, \textit{especially at higher noise levels}, even if a concomitant decrease in score was not evident due to our strict error criterion (small $\gamma$). Proceeding from Fig.~\ref{fig:b2}(b) to (c), increasing noise clearly led to responses resembling `random walks'.

Interestingly, \textit{at low noise levels}, a small injection of noise sometimes appeared to `kick' the model out of an incorrect answer and unexpectedly improved the score: note how a completely incorrect response for low noise in Fig.~\ref{fig:b2}(a) (its quadratic dispersion hardly traced) resolved to a less--incorrect response in (b) (the dispersion now approximately traced), before the aforementioned `random walk' set in at higher noise levels in (c) and severely degraded its responses. We postulate that a small, salutary dose of noise may nudge models away from using incorrect heuristics that short–circuit its reasoning. (The tendency of LLMs to resort to heuristics, noticed by a theoretical condensed matter benchmark, \citet{pan2025}, has also been explored by \citet{ma2025}, which distinguished this brittle form of memorization–based reasoning from genuine abstract reasoning.)

\section{Impact \& Limitations} \label{section5}

We have presented \textbf{CME–SpectrumBench}, the first benchmark testing how frontier LLMs analyze spectral scientific data in condensed matter \textbf{\textit{experiments}} (CME), a timely and essential counterpart to recent benchmarks in condensed matter \textbf{\textit{theory}} \cite{pan2025, wang2025a, wang2025}. CME–SpectrumBench tests how well models extract meaningful information from noisy experimental data, which is a \textbf{long–context inverse problem} that requires \textbf{dense context reasoning}. Models were scored by how accurately their numerical answers approached the ground truth, such that a difference equivalent to typical experimental standard deviation ($1\sigma$) is accorded a score of 0.5 out of 1.

\subsection{Agentic LLMs are now useful scientific coding assistants \ldots}

We showed that the best performing frontier models (as of May 2026) are competitive with handwritten code when questions were reframed as code generation problems, by enabling access to a \textbf{Python interpreter}, with/without \textbf{ReAct prompting}. LLMs could thus be useful for preliminary analysis, especially in processing large datasets where speed is important. A key limitation in scientific progress is data quality itself; sophisticated analysis cannot compensate for poor data. Agents working in real time could be valuable for analyzing large datasets quickly and suggesting experimental adjustments while data is being collected, such as during high–throughput experimental runs, rather than during post–processing (as in CME–SpectrumBench). In the immediate future, this points toward evaluating scientific agents in dynamic, real–world experimental scenarios, akin to $\tau$–bench \citep{yao2024}.

\subsection{\ldots  but progress is needed towards foundation models in physics}

Under \textbf{direct query}, models extracted scalar features (Tier I) more easily than they traced patterns (Tier II) or computed derived quantities (Tier III). Performance degraded with increased data resolution (Fig.~\ref{fig:ablations}) and noise/convolution levels (Figs.~\ref{fig:leaderboards}, \ref{fig:tiers}, and \ref{fig:b2}). We postulate that \textbf{worse performance at higher data resolution} (especially Tier II questions) is related to the difficulty of dense context reasoning \citep{kurkin2025}, extracting latent structures \citep{vodrahalli2024}, and/or a key drawback of the transformer architecture: \textit{that computational complexity scales quadratically with input token count}.

We surmise that the problem of long contexts is entangled with that of representation, because finding a more compact representation containing the same amount of relevant global information shortens the context. Noting the tension between representing fine–grained local structures and coarse–grained global structures, we anticipate that \textit{identifying good representations and architectures} will be a key challenge for building \textbf{long–context foundation models in physics} (\textit{which must ingest raw data rather than interface with them through code}), and that solutions to this problem will be varied. This prediction reflects the history of (far more advanced) biological foundation models. In an early example, $k$–mer tokenization in DNABERT \citep{ji2021} was replaced by byte–pair encoding in DNABERT–2 \citep{zhou2024} for computational efficiency. The multi–modal long–context foundation model Evo \citep{nguyen2024, brixi2026} adopted a StripedHyena model architecture specifically to balance \textit{token–level attention} of transformers with \textit{identifying patterns across long sequences} using convolutional operators. Lastly, ESM3 \citep{hayes2025} improved upon its predecessors by using a discrete encoder \citep{vdo2018} to represent continuously defined (3D) protein structures with discrete tokens, potentially useful for continuous data in physics.

\subsection{Limitations and further work}

We note several limitations of CME–SpectrumBench. \textbf{Firstly}, experimental datasets in ARPES and RIXS are sometimes \textit{more extensive} (such as 3D spectra consisting of stacks of 2D spectra) or possess \textit{much higher resolution} than evaluated here, which are limited by the context windows of today's frontier models. \textbf{Secondly}, \textit{fine–tuning or post–training} of models, although promising for improving scores and a rich avenue for future work, is outside the scope of our three–shot, closed–book evaluation. \textit{Vision language models} were not tested: the sacrifice in quantitative accuracy due to their patch embedding (an effective coarse–graining), arbitrariness of color scales to represent intensities (whose raw numerical values are important), and incompatibility with future extensions to 3D spectral data or non–Cartesian binning in real data point to an evaluative \textit{cul–de–sac}. \textbf{Thirdly}, ARPES and RIXS are \textit{a small fraction of spectroscopies} in physics and chemistry. Nevertheless, they are \textit{representative}: we stress that noise and convolution, the key attributes of ARPES and RIXS complicating the inverse problem, recur for many modalities of data defined over continuous domains — the form that predominates in spectroscopies. For example, astronomical images are corrupted by Poisson noise \citep{shamshad2018} convolved with point spread functions such as the characteristic speckles of the James Webb Space Telescope \citep{kinakh2024}; measuring jets in high–energy physics involves sifting through background noise \citep{sjo2012} and accounting for calorimetric instrumental resolution \citep{lobban2002}. \textbf{Lastly}, ARPES and RIXS are merely two, \textit{nonetheless crucial} techniques in the broad field of condensed matter, \textit{of which many others are non–spectroscopic}, and whose data future multi–modal foundation models must be able to comprehend natively. To this end, we wish to welcome scientific collaborators to extend our benchmark to encompass other experimental techniques. Because experts in techniques who implement these extensions may adopt approaches different from us, we have formatted questions in CME–SpectrumBench in a general yet standardized manner to ensure consistent application and analysis, while allowing freedom in realizing details \hyperref[appendixC]{(Appendix C)}.

\begin{ack}
This research benefitted from resources of the Stanford Institute for Materials and Energy Sciences, supported by the DOE Office of Basic Energy Sciences, Division of Materials Sciences and Engineering under contract no. DE-AC02-76SF00515. We thank Chunjing Jia, Karen Lei, Sebastien Abadi, and Bo Liu for helpful comments and guidance.
\end{ack}

\bibliography{bibliography}

@article{sobota2021,
  title = {Angle-resolved photoemission studies of quantum materials},
  author = {Sobota, Jonathan A. and He, Yu and Shen, Zhi-Xun},
  journal = {Rev. Mod. Phys.},
  volume = {93},
  issue = {2},
  pages = {025006},
  year = {2021},
  doi = {10.1103/RevModPhys.93.025006},
  url = {https://link.aps.org/doi/10.1103/RevModPhys.93.025006}
}

@article{hendrycks2021,
      title={{M}easuring {M}assive {M}ultitask {L}anguage {U}nderstanding}, 
      author={Dan Hendrycks and Collin Burns and Steven Basart and Andy Zou and Mantas Mazeika and Dawn Song and Jacob Steinhardt},
      year={2021},
      journal={Proceedings of the 9th International Conference on Learning Representations (ICLR)},
      url={https://arxiv.org/abs/2009.03300}, 
}

@misc{cobbe2021,
      title={{T}raining {V}erifiers to {S}olve {M}ath {W}ord {P}roblems}, 
      author={Karl Cobbe and Vineet Kosaraju and Mohammad Bavarian and Mark Chen and Heewoo Jun and Lukasz Kaiser and Matthias Plappert and Jerry Tworek and Jacob Hilton and Reiichiro Nakano and Christopher Hesse and John Schulman},
      year={2021},
      eprint={2110.14168},
      archivePrefix={arXiv},
      primaryClass={cs.LG},
      url={https://arxiv.org/abs/2110.14168}, 
}

@misc{glazer2024,
      title={{F}rontier{M}ath: {A} {B}enchmark for {E}valuating {A}dvanced {M}athematical {R}easoning in {AI}}, 
      author={Elliot Glazer and Ege Erdil and Tamay Besiroglu and Diego Chicharro and Evan Chen and Alex Gunning and Caroline Falkman Olsson and Jean-Stanislas Denain and Anson Ho and Emily de Oliveira Santos and Olli Järviniemi and Matthew Barnett and Robert Sandler and Matej Vrzala and Jaime Sevilla and Qiuyu Ren and Elizabeth Pratt and Lionel Levine and Grant Barkley and Natalie Stewart and Bogdan Grechuk and Tetiana Grechuk and Shreepranav Varma Enugandla and Mark Wildon},
      year={2024},
      eprint={2411.04872},
      archivePrefix={arXiv},
      primaryClass={cs.AI},
      url={https://arxiv.org/abs/2411.04872}, 
}

@misc{phan2025,
      title={{H}umanity's {L}ast {E}xam}, 
      author={Long Phan and Dan Hendrycks et al.},
      year={2025},
      eprint={2501.14249},
      archivePrefix={arXiv},
      primaryClass={cs.LG},
      url={https://arxiv.org/abs/2501.14249}, 
}

@misc{hua2025,
      title={{I}nduction{B}ench: {LLM}s {F}ail in the {S}implest {C}omplexity {C}lass}, 
      author={Wenyue Hua and Tyler Wong and Sun Fei and Liangming Pan and Adam Jardine and William Yang Wang},
      year={2025},
      eprint={2502.15823},
      archivePrefix={arXiv},
      primaryClass={cs.LG},
      url={https://arxiv.org/abs/2502.15823}, 
}

@misc{ravichander2021,
      title={{N}oise{QA}: {C}hallenge {S}et {E}valuation for {U}ser-{C}entric {Q}uestion {A}nswering}, 
      author={Abhilasha Ravichander and Siddharth Dalmia and Maria Ryskina and Florian Metze and Eduard Hovy and Alan W Black},
      year={2021},
      eprint={2102.08345},
      archivePrefix={arXiv},
      primaryClass={cs.CL},
      url={https://arxiv.org/abs/2102.08345}, 
}

@article{tay2020,
      title={{L}ong {R}ange {A}rena: {A} {B}enchmark for {E}fficient {T}ransformers}, 
      author={Yi Tay and Mostafa Dehghani and Samira Abnar and Yikang Shen and Dara Bahri and Philip Pham and Jinfeng Rao and Liu Yang and Sebastian Ruder and Donald Metzler},
      year={2020},
      journal={Proceedings of the 9th International Conference on Learning Representations (ICLR)},
      url={https://arxiv.org/abs/2011.04006}, 
}

@misc{vodrahalli2024,
      title={{M}ichelangelo: {L}ong {C}ontext {E}valuations {B}eyond {H}aystacks via {L}atent {S}tructure {Q}ueries}, 
      author={Kiran Vodrahalli and Santiago Ontanon and Nilesh Tripuraneni and Kelvin Xu and Sanil Jain and Rakesh Shivanna and Jeffrey Hui and Nishanth Dikkala and Mehran Kazemi and Bahare Fatemi and Rohan Anil and Ethan Dyer and Siamak Shakeri and Roopali Vij and Harsh Mehta and Vinay Ramasesh and Quoc Le and Ed Chi and Yifeng Lu and Orhan Firat and Angeliki Lazaridou and Jean-Baptiste Lespiau and Nithya Attaluri and Kate Olszewska},
      year={2024},
      eprint={2409.12640},
      archivePrefix={arXiv},
      primaryClass={cs.CL},
      url={https://arxiv.org/abs/2409.12640}, 
}

@misc{chollet2025,
      title={{ARC} {P}rize 2024: {T}echnical {R}eport}, 
      author={Francois Chollet and Mike Knoop and Gregory Kamradt and Bryan Landers},
      year={2025},
      eprint={2412.04604},
      archivePrefix={arXiv},
      primaryClass={cs.AI},
      url={https://arxiv.org/abs/2412.04604}, 
}

@article{yao2024,
      title={$\tau$-bench: {A} {B}enchmark for {T}ool-{A}gent-{U}ser {I}nteraction in {R}eal-{W}orld {D}omains}, 
      author={Shunyu Yao and Noah Shinn and Pedram Razavi and Karthik Narasimhan},
      year={2025},
      journal={Proceedings of the 13th International Conference on Learning Representations (ICLR)},
      url={https://arxiv.org/abs/2406.12045}, 
}

@misc{kaplan2020,
      title={{S}caling {L}aws for {N}eural {L}anguage {M}odels}, 
      author={Jared Kaplan and Sam McCandlish and Tom Henighan and Tom B. Brown and Benjamin Chess and Rewon Child and Scott Gray and Alec Radford and Jeffrey Wu and Dario Amodei},
      year={2020},
      eprint={2001.08361},
      archivePrefix={arXiv},
      primaryClass={cs.LG},
      url={https://arxiv.org/abs/2001.08361}, 
}

@article{turing50,
    author = {Turing, Alan},
    journal = {Mind},
    volume = {LIX},
    issue = {236},
    pages = {433--460},
    title = {{C}omputing machinery and intelligence},
    year = {1950},
    url = {https://doi.org/10.1093/mind/LIX.236.433}
}

@ARTICLE{deng2012,
    author={Deng, Li},
    journal={IEEE Signal Processing Magazine}, 
    title={The {MNIST} Database of {H}andwritten {D}igit {I}mages for {M}achine {L}earning {R}esearch [{B}est of the {W}eb]}, 
    year={2012},
    volume={29},
    number={6},
    pages={141-142},
    doi={10.1109/MSP.2012.2211477},
    url={https://ieeexplore.ieee.org/document/6296535}}

@INPROCEEDINGS{deng2009,
    author={Deng, Jia and Dong, Wei and Socher, Richard and Li, Li-Jia and Kai Li and Li Fei-Fei},
    booktitle={2009 IEEE Conference on Computer Vision and Pattern Recognition (CVPR)}, 
    title={{I}mage{N}et: {A} large-scale hierarchical image database}, 
    year={2009},
    volume={},
    number={},
    pages={248-255},
    doi={10.1109/CVPR.2009.5206848},
    url={https://ieeexplore.ieee.org/document/5206848}}

@misc{kelly2023,
      title={The {UCI} {M}achine {L}earning {R}epository}, 
      author={Markelle Kelly and Rachel Longjohn and Kolby Nottingham},
      year={2023},
      url={https://archive.ics.uci.edu}
}

@inproceedings{rajpurkar2016,
    title = "{SQ}u{AD}: 100,000+ {Q}uestions for {M}achine {C}omprehension of {T}ext",
    author = "Rajpurkar, Pranav  and
      Zhang, Jian  and
      Lopyrev, Konstantin  and
      Liang, Percy",
    booktitle = "Proceedings of the 2016 Conference on Empirical Methods in Natural Language Processing (EMNLP)",
    year = "2016",
    doi = "10.18653/v1/D16-1264",
    pages = "2383--2392",
    url = "https://aclanthology.org/D16-1264/"
}

@inproceedings{wang2018,
    title = "{GLUE}: A {M}ulti-{T}ask {B}enchmark and {A}nalysis {P}latform for {N}atural {L}anguage {U}nderstanding",
    author = "Wang, Alex  and
      Singh, Amanpreet  and
      Michael, Julian  and
      Hill, Felix  and
      Levy, Omer  and
      Bowman, Samuel",
    booktitle = "Proceedings of the 2018 {EMNLP} Workshop {B}lackbox{NLP}: Analyzing and Interpreting Neural Networks for {NLP}",
    year = "2018",
    doi = "10.18653/v1/W18-5446",
    pages = "353--355",
    url = "https://aclanthology.org/W18-5446/"
}

@inbook{wang2019,
author = {Wang, Alex and Pruksachatkun, Yada and Nangia, Nikita and Singh, Amanpreet and Michael, Julian and Hill, Felix and Levy, Omer and Bowman, Samuel R.},
title = {{S}uper{GLUE}: a stickier benchmark for general-purpose language understanding systems},
year = {2019},
booktitle = {Proceedings of the 33rd International Conference on Neural Information Processing Systems (NIPS)},
articleno = {294},
numpages = {15},
url = {https://dl.acm.org/doi/10.5555/3454287.3454581}
}

@inproceedings{papineni2002,
    title = "{B}leu: a {M}ethod for {A}utomatic {E}valuation of {M}achine {T}ranslation",
    author = "Papineni, Kishore  and
      Roukos, Salim  and
      Ward, Todd  and
      Zhu, Wei-Jing",
    booktitle = "Proceedings of the 40th Annual Meeting of the Association for Computational Linguistics",
    year = "2002",
    doi = "10.3115/1073083.1073135",
    pages = "311--318",
    url = "https://aclanthology.org/P02-1040/"
}

@inproceedings{chiang2024,
author = {Chiang, Wei-Lin and Zheng, Lianmin and Sheng, Ying and Angelopoulos, Anastasios N. and Li, Tianle and Li, Dacheng and Zhu, Banghua and Zhang, Hao and Jordan, Michael I. and Gonzalez, Joseph E. and Stoica, Ion},
title = {{C}hatbot arena: an open platform for evaluating {LLM}s by human preference},
year = {2024},
booktitle = {Proceedings of the 41st International Conference on Machine Learning (ICML)},
articleno = {331},
numpages = {30},
url = {https://dl.acm.org/doi/abs/10.5555/3692070.3692401}
}

@article{zhao2025,
      title={{S}ci{A}rena: {A}n {O}pen {E}valuation {P}latform for {F}oundation {M}odels in {S}cientific {L}iterature {T}asks}, 
      author={Yilun Zhao and Kaiyan Zhang and Tiansheng Hu and Sihong Wu and Ronan Le Bras and Taira Anderson and Jonathan Bragg and Joseph Chee Chang and Jesse Dodge and Matt Latzke and Yixin Liu and Charles McGrady and Xiangru Tang and Zihang Wang and Chen Zhao and Hannaneh Hajishirzi and Doug Downey and Arman Cohan},
      journal={Proceedings of the 39th International Conference on Neural Information Processing Systems (NIPS)}, 
      url={https://arxiv.org/abs/2507.01001}, 
      year={2025}
}

@article{liang2023,
      title={{H}olistic {E}valuation of {L}anguage {M}odels}, 
      author={Percy Liang and Rishi Bommasani and Tony Lee and Dimitris Tsipras and Dilara Soylu and Michihiro Yasunaga and Yian Zhang and Deepak Narayanan and Yuhuai Wu and Ananya Kumar and Benjamin Newman and Binhang Yuan and Bobby Yan and Ce Zhang and Christian Cosgrove and Christopher D. Manning and Christopher Ré and Diana Acosta-Navas and Drew A. Hudson and Eric Zelikman and Esin Durmus and Faisal Ladhak and Frieda Rong and Hongyu Ren and Huaxiu Yao and Jue Wang and Keshav Santhanam and Laurel Orr and Lucia Zheng and Mert Yuksekgonul and Mirac Suzgun and Nathan Kim and Neel Guha and Niladri Chatterji and Omar Khattab and Peter Henderson and Qian Huang and Ryan Chi and Sang Michael Xie and Shibani Santurkar and Surya Ganguli and Tatsunori Hashimoto and Thomas Icard and Tianyi Zhang and Vishrav Chaudhary and William Wang and Xuechen Li and Yifan Mai and Yuhui Zhang and Yuta Koreeda},
      year={2023},
      journal={Transactions on Machine Learning Research (TMLR)},
      url={https://arxiv.org/abs/2211.09110}, 
}

@misc{dubois2025,
      title={{L}ength-{C}ontrolled {A}lpaca{E}val: {A} {S}imple {W}ay to {D}ebias {A}utomatic {E}valuators}, 
      author={Yann Dubois and Balázs Galambosi and Percy Liang and Tatsunori B. Hashimoto},
      year={2025},
      eprint={2404.04475},
      archivePrefix={arXiv},
      primaryClass={cs.LG},
      url={https://arxiv.org/abs/2404.04475},
}

@article{lin2024,
      title={{W}ild{B}ench: {B}enchmarking {LLM}s with {C}hallenging {T}asks from {R}eal {U}sers in the {W}ild}, 
      author={Bill Yuchen Lin and Yuntian Deng and Khyathi Chandu and Faeze Brahman and Abhilasha Ravichander and Valentina Pyatkin and Nouha Dziri and Ronan Le Bras and Yejin Choi},
      year={2025},
      journal={Proceedings of the 13th International Conference on Learning Representations (ICLR)},
      url={https://arxiv.org/abs/2406.04770}, 
}

@article{white2025,
      title={{L}ive{B}ench: {A} {C}hallenging, {C}ontamination-{L}imited {LLM} {B}enchmark}, 
      author={Colin White and Samuel Dooley and Manley Roberts and Arka Pal and Ben Feuer and Siddhartha Jain and Ravid Shwartz-Ziv and Neel Jain and Khalid Saifullah and Sreemanti Dey and Shubh-Agrawal and Sandeep Singh Sandha and Siddartha Naidu and Chinmay Hegde and Yann LeCun and Tom Goldstein and Willie Neiswanger and Micah Goldblum},
      year={2025},
      journal={Proceedings of the 13th International Conference on Learning Representations (ICLR)},
      url={https://arxiv.org/abs/2406.19314}, 
}

@misc{rein2023,
      title={{GPQA}: {A} {G}raduate-{L}evel {G}oogle-{P}roof {Q\&A} {B}enchmark}, 
      author={David Rein and Betty Li Hou and Asa Cooper Stickland and Jackson Petty and Richard Yuanzhe Pang and Julien Dirani and Julian Michael and Samuel R. Bowman},
      year={2023},
      eprint={2311.12022},
      archivePrefix={arXiv},
      primaryClass={cs.AI},
      url={https://arxiv.org/abs/2311.12022}, 
}

@article{jimenez2024,
      title={{SWE}-bench: {C}an {L}anguage {M}odels {R}esolve {R}eal-{W}orld {G}itHub {I}ssues?}, 
      author={Carlos E. Jimenez and John Yang and Alexander Wettig and Shunyu Yao and Kexin Pei and Ofir Press and Karthik Narasimhan},
      year={2024},
      journal={Proceedings of the 12th International Conference on Learning Representations (ICLR)},
      url={https://arxiv.org/abs/2310.06770}, 
}

@article{chung2025,
      title={{T}heoretical {P}hysics {B}enchmark ({TPB}ench) – a {D}ataset and {S}tudy of {AI} {R}easoning {C}apabilities in {T}heoretical {P}hysics}, 
      author={Daniel J. H. Chung and Zhiqi Gao and Yurii Kvasiuk and Tianyi Li and Moritz Münchmeyer and Maja Rudolph and Frederic Sala and Sai Chaitanya Tadepalli},
      year={2025},
      journal={Machine Learning: Science and Technology},
      volume = {6},
      number = {3},
      pages = {030505},
      url={https://iopscience.iop.org/article/10.1088/2632-2153/adfcb0}, 
}

@misc{wang2025,
      title={{QMBench: A Research Level Benchmark for Quantum Materials Research}}, 
      author={Yanzhen Wang and Yiyang Jiang and Diana Golovanova and Kamal Das and Hyeonhu Bae and Yufei Zhao and Huu-Thong Le and Abhinava Chatterjee and Yunzhe Liu and Chao-Xing Liu and Felipe H. da Jornada and Binghai Yan and Xiao-Liang Qi},
      year={2025},
      eprint={2512.19753},
      archivePrefix={arXiv},
      primaryClass={cond-mat.mtrl-sci},
      url={https://arxiv.org/abs/2512.19753}, 
}

@inbook{stanley1971,
author = {H. E. Stanley},
title = {{I}ntroduction to {P}hase {T}ransitions and {C}ritical {P}henomena},
year = {1971},
booktitle = {Oxford University Press}
}

@article{damascelli2003,
  title = {Angle-resolved photoemission studies of the cuprate superconductors},
  author = {Damascelli, Andrea and Hussain, Zahid and Shen, Zhi-Xun},
  journal = {Rev. Mod. Phys.},
  volume = {75},
  issue = {2},
  pages = {473--541},
  year = {2003},
  publisher = {American Physical Society},
  doi = {10.1103/RevModPhys.75.473},
  url = {https://link.aps.org/doi/10.1103/RevModPhys.75.473}
}

@article{liu2024,
    title = "{L}ost in the {M}iddle: {H}ow {L}anguage {M}odels {U}se {L}ong {C}ontexts",
    author = "Liu, Nelson F.  and
      Lin, Kevin  and
      Hewitt, John  and
      Paranjape, Ashwin  and
      Bevilacqua, Michele  and
      Petroni, Fabio  and
      Liang, Percy",
    journal = "Transactions of the Association for Computational Linguistics",
    volume = "12",
    year = "2024",
    url = "https://aclanthology.org/2024.tacl-1.9/",
    doi = "10.1162/tacl_a_00638",
    pages = "157--173"
}

@article{templeton2024,
    title={{Scaling Monosemanticity: Extracting Interpretable Features from Claude 3 Sonnet}},
    author={Templeton, Adly and Conerly, Tom and Marcus, Jonathan and Lindsey, Jack and Bricken, Trenton and Chen, Brian and Pearce, Adam and Citro, Craig and Ameisen, Emmanuel and Jones, Andy and Cunningham, Hoagy and Turner, Nicholas L and McDougall, Callum and MacDiarmid, Monte and Freeman, C. Daniel and Sumers, Theodore R. and Rees, Edward and Batson, Joshua and Jermyn, Adam and Carter, Shan and Olah, Chris and Henighan, Tom},
    year={2024},
    journal={Transformer Circuits Thread},
       url={https://transformer-circuits.pub/2024/scaling-monosemanticity/index.html}
}

@article{nguyen2024,
author = {Eric Nguyen  and Michael Poli  and Matthew G. Durrant  and Brian Kang  and Dhruva Katrekar  and David B. Li  and Liam J. Bartie  and Armin W. Thomas  and Samuel H. King  and Garyk Brixi  and Jeremy Sullivan  and Madelena Y. Ng  and Ashley Lewis  and Aaron Lou  and Stefano Ermon  and Stephen A. Baccus  and Tina Hernandez-Boussard  and Christopher Ré  and Patrick D. Hsu  and Brian L. Hie },
title = {Sequence modeling and design from molecular to genome scale with {E}vo},
journal = {Science},
volume = {386},
number = {6723},
pages = {eado9336},
year = {2024},
doi = {10.1126/science.ado9336},
URL = {https://www.science.org/doi/abs/10.1126/science.ado9336},
eprint = {https://www.science.org/doi/pdf/10.1126/science.ado9336}
}

@article{kim2021,
author = {Kim, Hyunseob and Lee, Jeongcheol and Ahn, Sunil and Lee, Jongsuk Ruth},
title = {A merged molecular representation learning for molecular properties prediction with a web-based service},
journal = {Scientific Reports},
volume = {11},
number = {1},
pages = {11028},
year = {2021},
doi = {10.1038/s41598-021-90259-7},
URL = {https://doi.org/10.1038/s41598-021-90259-7}
}

@misc{chithrananda2020,
      title={{ChemBERTa: Large-Scale Self-Supervised Pretraining for Molecular Property Prediction}}, 
      author={Seyone Chithrananda and Gabriel Grand and Bharath Ramsundar},
      year={2020},
      eprint={2010.09885},
      archivePrefix={arXiv},
      primaryClass={cs.LG},
      url={https://arxiv.org/abs/2010.09885}, 
}

@article{zhang2025,
author = {Zhang, Yanbo and Khan, Sumeer A. and Mahmud, Adnan and Yang, Huck and Lavin, Alexander and Levin, Michael and Frey, Jeremy and Dunnmon, Jared and Evans, James and Bundy, Alan and Dzeroski, Saso and Tegner, Jesper and Zenil, Hector},
title = {Exploring the role of large language models in the scientific method: from hypothesis to discovery},
journal = {npj Artificial Intelligence},
volume = {1},
number = {1},
pages = {14},
year = {2025},
doi = {10.1038/s44387-025-00019-5},
URL = {https://doi.org/10.1038/s44387-025-00019-5}
}

@article{liang2025,
author = {Liang, Weixin and Zhang, Yaohui and Wu, Zhengxuan and Lepp, Haley and Ji, Wenlong and Zhao, Xuandong and Cao, Hancheng and Liu, Sheng and He, Siyu and Huang, Zhi and Yang, Diyi and Potts, Christopher and Manning, Christopher D. and Zou, James},
title = {Quantifying large language model usage in scientific papers},
journal = {Nature Human Behaviour},
year = {2025},
doi = {10.1038/s41562-025-02273-8},
URL = {https://doi.org/10.1038/s41562-025-02273-8}
}

@article{chen2024,
  title={{ScienceAgentBench: Toward Rigorous Assessment of Language Agents for Data-Driven Scientific Discovery}}, 
  author={Ziru Chen and Shijie Chen and Yuting Ning and Qianheng Zhang and Boshi Wang and Botao Yu and Yifei Li and Zeyi Liao and Chen Wei and Zitong Lu and Vishal Dey and Mingyi Xue and Frazier N. Baker and Benjamin Burns and Daniel Adu-Ampratwum and Xuhui Huang and Xia Ning and Song Gao and Yu Su and Huan Sun},
  year={2025},
  journal={Proceedings of the 13th International Conference on Learning Representations (ICLR)},
  url={https://iclr.cc/virtual/2025/poster/32108}
  }

@article{majumder2024,
      title={{DiscoveryBench: Towards Data-Driven Discovery with Large Language Models}}, 
      author={Bodhisattwa Prasad Majumder and Harshit Surana and Dhruv Agarwal and Bhavana Dalvi Mishra and Abhijeetsingh Meena and Aryan Prakhar and Tirth Vora and Tushar Khot and Ashish Sabharwal and Peter Clark},
      year={2025},
      journal={Proceedings of the 13th International Conference on Learning Representations (ICLR)},
      url={https://arxiv.org/abs/2407.01725}, 
}

@misc{jansen2025,
      title={{CodeScientist: End-to-End Semi-Automated Scientific Discovery with Code-based Experimentation}}, 
      author={Peter Jansen and Oyvind Tafjord and Marissa Radensky and Pao Siangliulue and Tom Hope and Bhavana Dalvi Mishra and Bodhisattwa Prasad Majumder and Daniel S. Weld and Peter Clark},
      year={2025},
      eprint={2503.22708},
      archivePrefix={arXiv},
      primaryClass={cs.AI},
      url={https://arxiv.org/abs/2503.22708}, 
}

@misc{lu2024,
      title={{The AI Scientist: Towards Fully Automated Open-Ended Scientific Discovery}}, 
      author={Chris Lu and Cong Lu and Robert Tjarko Lange and Jakob Foerster and Jeff Clune and David Ha},
      year={2024},
      eprint={2408.06292},
      archivePrefix={arXiv},
      primaryClass={cs.AI},
      url={https://arxiv.org/abs/2408.06292}, 
}

@article{yamada2025,
      title={{The AI Scientist-v2: Workshop-Level Automated Scientific Discovery via Agentic Tree Search}}, 
      author={Yutaro Yamada and Robert Tjarko Lange and Cong Lu and Shengran Hu and Chris Lu and Jakob Foerster and Jeff Clune and David Ha},
      year={2025},
      journal={Proceedings of the 13th International Conference on Learning Representations (ICLR)},
      url={https://arxiv.org/abs/2504.08066}, 
}

@article{polanyi1962,
title={{The Republic of Science}}, 
author={Polanyi, Michael},
year={1962},
journal = {Minerva},
volume = {1},
number = {1},
pages = {54—73},
doi = {10.1007/BF01101453},
url={https://doi.org/10.1007/BF01101453}, 
}

@misc{liu2024a,
title={{Are LLMs Capable of Data-based Statistical and Causal Reasoning? Benchmarking Advanced Quantitative Reasoning with Data}}, 
author={Xiao Liu and Zirui Wu and Xueqing Wu and Pan Lu and Kai-Wei Chang and Yansong Feng},
year={2024},
eprint={2402.17644},
archivePrefix={arXiv},
primaryClass={cs.CL},
url={https://arxiv.org/abs/2402.17644}, 
}

@article{shamshad2018,
author = {Shamshad, Fahad and Riaz, M. Mohsin and Ghafoor, Abdul},
title = {{Poisson Denoising for Astronomical Images}},
journal = {Advances in Astronomy},
volume = {2018},
number = {1},
pages = {2417939},
doi = {https://doi.org/10.1155/2018/2417939},
url = {https://onlinelibrary.wiley.com/doi/abs/10.1155/2018/2417939},
eprint = {https://onlinelibrary.wiley.com/doi/pdf/10.1155/2018/2417939},
year = {2018}
}

@Article{kinakh2024,
AUTHOR = {Kinakh, Vitaliy and Belousov, Yury and Quétant, Guillaume and Drozdova, Mariia and Holotyak, Taras and Schaerer, Daniel and Voloshynovskiy, Slava},
TITLE = {Hubble Meets Webb: Image-to-Image Translation in Astronomy},
JOURNAL = {Sensors},
VOLUME = {24},
YEAR = {2024},
NUMBER = {4},
ARTICLE-NUMBER = {1151},
URL = {https://www.mdpi.com/1424-8220/24/4/1151},
PubMedID = {38400307},
ISSN = {1424-8220},
DOI = {10.3390/s24041151}
}

@article{baldi2016,
  title = {Jet substructure classification in high-energy physics with deep neural networks},
  author = {Baldi, Pierre and Bauer, Kevin and Eng, Clara and Sadowski, Peter and Whiteson, Daniel},
  journal = {Phys. Rev. D},
  volume = {93},
  issue = {9},
  pages = {094034},
  numpages = {8},
  year = {2016},
  month = {May},
  publisher = {American Physical Society},
  doi = {10.1103/PhysRevD.93.094034},
  url = {https://link.aps.org/doi/10.1103/PhysRevD.93.094034}
}

@techreport{sjo2012,
      author        = "Sjölin, J",
      title         = "{Estimation Techniques for Instrumental Backgrounds at the
                       LHC}",
      institution   = "CERN",
      reportNumber  = "ATL-PHYS-PROC-2012-007",
      address       = "Geneva",
      year          = "2012",
      url           = "https://cds.cern.ch/record/1415690",
}

@article{lobban2002,
title = {On the energy measurement of hadron jets},
journal = {Nuclear Instruments and Methods in Physics Research Section A: Accelerators, Spectrometers, Detectors and Associated Equipment},
volume = {495},
number = {2},
pages = {107-120},
year = {2002},
issn = {0168-9002},
doi = {https://doi.org/10.1016/S0168-9002(02)01615-7},
url = {https://www.sciencedirect.com/science/article/pii/S0168900202016157},
author = {Olga Lobban and Aravindhan Sriharan and Richard Wigmans}
}

@article{kim2021a,
    author = {Kim, Younsik and Oh, Dongjin and Huh, Soonsang and Song, Dongjoon and Jeong, Sunbeom and Kwon, Junyoung and Kim, Minsoo and Kim, Donghan and Ryu, Hanyoung and Jung, Jongkeun and Kyung, Wonshik and Sohn, Byungmin and Lee, Suyoung and Hyun, Jounghoon and Lee, Yeonghoon and Kim, Yeongkwan and Kim, Changyoung},
    title = {Deep learning-based statistical noise reduction for multidimensional spectral data},
    journal = {Review of Scientific Instruments},
    volume = {92},
    number = {7},
    pages = {073901},
    year = {2021},
    month = {07},
    issn = {0034-6748},
    doi = {10.1063/5.0054920},
    url = {https://doi.org/10.1063/5.0054920},
    eprint = {https://pubs.aip.org/aip/rsi/article-pdf/doi/10.1063/5.0054920/16017333/073901_1_online.pdf}
}

@techreport{genesis2025,
  Title                    = {{Sections 3(iv, v) and 4(iii, v), Launching the Genesis Mission, Executive Order}},
  Date                     = {2025-11-24},
  Author                   = {Trump, Donald J},
  Institution              = {The White House},
  Type                     = {Executive Order},
  Year                     = {2025},
url = {https://www.whitehouse.gov/presidential-actions/2025/11/launching-the-genesis-mission/}
}

@article{yang2018,
    author = {Yang, Haifeng and Liang, Aiji and Chen, Cheng and Zhang, Chaofan and Schroeter, Niels B. M. and Chen, Yulin},
    title = {Visualizing electronic structures of quantum materials by angle-resolved photoemission spectroscopy},
    journal = {Nature Reviews Materials},
    volume = {3},
    number = {9},
    pages = {341–353},
    year = {2018},
    doi = {10.1038/s41578-018-0047-2},
    url = {https://doi.org/10.1038/s41578-018-0047-2}
}

@article{goyal2025,
    author = {Goyal, Rajat Kumar and Maharaj, Shivam and Kumar, Pawan and Chandrasekhar, M.},
    title = {Exploring quantum materials and applications: a review},
    journal = {Journal of Materials Science: Materials in Engineering},
    volume = {20},
    number = {1},
    pages = {4},
    year = {2025},
    doi = {10.1186/s40712-024-00202-7},
    url = {https://doi.org/10.1186/s40712-024-00202-7}
}

@article{ying2022,
title = {Solving inverse problems with deep learning},
journal = {International Congress of Mathematicians},
author={Ying, Lexing},
volume = {VII},
pages={5154–5175},
publisher = {Euroopean Mathematical Society},
year = {2022},
doi = {10.4171/ICM2022/49},
URL = {https://ems.press/books/standalone/279/5595}
}

@article{lecun2022,
title = {A Path Towards Autonomous Machine Intelligence},
journal = {OpenReview},
year = {2022},
url = {https://openreview.net/forum?id=BZ5a1r-kVsf},
author = {LeCun, Yann}
}

@misc{pan2025,
      title={{CMT-Benchmark: A Benchmark for Condensed Matter Theory Built by Expert Researchers}}, 
      author={Haining Pan and James V. Roggeveen and Erez Berg and Juan Carrasquilla and Debanjan Chowdhury and Surya Ganguli and Federico Ghimenti and Juraj Hasik and Henry Hunt and Hong-Chen Jiang and Mason Kamb and Ying-Jer Kao and Ehsan Khatami and Michael J. Lawler and Di Luo and Titus Neupert and Xiaoliang Qi and Michael P. Brenner and Eun-Ah Kim},
      year={2025},
      url={https://openreview.net/forum?id=pX6B28ynNh}, 
}

@misc{wang2025a,
      title={{CMPhysBench: A Benchmark for Evaluating Large Language Models in Condensed Matter Physics}}, 
      author={Weida Wang and Dongchen Huang and Jiatong Li and Tengchao Yang and Ziyang Zheng and Di Zhang and Dong Han and Benteng Chen and Binzhao Luo and Zhiyu Liu and Kunling Liu and Zhiyuan Gao and Shiqi Geng and Wei Ma and Jiaming Su and Xin Li and Shuchen Pu and Yuhan Shui and Qianjia Cheng and Zhihao Dou and Dongfei Cui and Changyong He and Jin Zeng and Zeke Xie and Mao Su and Dongzhan Zhou and Yuqiang Li and Wanli Ouyang and Yunqi Cai and Xi Dai and Shufei Zhang and Lei Bai and Jinguang Cheng and Zhong Fang and Hongming Weng},
      year={2025},
      url={https://openreview.net/forum?id=3d0FRYx0D0}, 
}

@article{ament2011,
  title = {{Resonant Inelastic X-ray Scattering Studies of Elementary Excitations}},
  author = {Ament, Luuk J. P. and van Veenendaal, Michel and Devereaux, Thomas P. and Hill, John P. and van den Brink, Jeroen},
  journal = {Rev. Mod. Phys.},
  volume = {83},
  issue = {2},
  pages = {705--767},
  year = {2011},
  doi = {10.1103/RevModPhys.83.705},
  url = {https://link.aps.org/doi/10.1103/RevModPhys.83.705}
}

@article{mitrano2024,
  title = {{Exploring Quantum Materials with Resonant Inelastic X-Ray Scattering}},
  author = {Mitrano, M. and Johnston, S. and Kim, Young-June and Dean, M. P. M.},
  journal = {Phys. Rev. X},
  volume = {14},
  issue = {4},
  pages = {040501},
  year = {2024},
  doi = {10.1103/PhysRevX.14.040501},
  url = {https://link.aps.org/doi/10.1103/PhysRevX.14.040501}
}

@misc{wang2025b,
      title={{Unable to Forget: Proactive Interference Reveals Working Memory Limits in LLMs Beyond Context Length}}, 
      author={{Chupei Wang and Jiaqiu Vince Sun}},
      year={2025},
      eprint={2506.08184},
      archivePrefix={arXiv},
      primaryClass={cs.CL},
      url={https://arxiv.org/abs/2506.08184}, 
}

@article{zhou2025,
      title={{GSM-Infinite: How Do Your LLMs Behave over Infinitely Increasing Context Length and Reasoning Complexity?}}, 
      author={Yang Zhou and Hongyi Liu and Zhuoming Chen and Yuandong Tian and Beidi Chen},
      year={2025},
      journal={Proceedings of the 42nd International Conference on Machine Learning (ICML)},
      url={https://arxiv.org/abs/2502.05252}, 
}

@misc{openai2025,
      title={{MRCR: Long context multiple needle in a haystack benchmark}}, 
      author={OpenAI},
      year={2025},
      url={https://huggingface.co/datasets/openai/mrcr}, 
}

@article{yao2023,
      title={{ReAct: Synergizing Reasoning and Acting in Language Models}}, 
      author={Shunyu Yao and Jeffrey Zhao and Dian Yu and Nan Du and Izhak Shafran and Karthik Narasimhan and Yuan Cao},
      year={2023},
      journal={Proceedings of the 11th International Conference on Learning Representations (ICLR)},
      url={https://arxiv.org/abs/2210.03629}, 
}

@misc{kurkin2025,
      title={{MMReD: a Cross-Modal Benchmark for Dense Context Reasoning}}, 
      author={Maxim Kurkin and Boris Shirokikh and Irina Abdullaeva and Viktoriya Chekalina},
      year={2025},
      url={https://openreview.net/forum?id=H6fM44DOHP}, 
}

@article{ma2025,
      title={{Benchmarking Abstract and Reasoning Abilities Through A Theoretical Perspective}}, 
      author={Qingchuan Ma and Yuhang Wu and Xiawu Zheng and Rongrong Ji},
      year={2025},
      journal={Proceedings of the 42nd International Conference on Machine Learning (ICML)},
      url={https://arxiv.org/abs/2505.23833}, 
}

@article{kuratov2024,
      title={{BABILong: Testing the Limits of LLMs with Long Context Reasoning-in-a-Haystack}}, 
      author={Yuri Kuratov and Aydar Bulatov and Petr Anokhin and Ivan Rodkin and Dmitry Sorokin and Artyom Sorokin and Mikhail Burtsev},
      year={2024},
      journal={Proceedings of the 38th International Conference on Neural Information Processing Systems (NIPS)},
      url={https://arxiv.org/abs/2406.10149}, 
}

@misc{hsieh2024,
      title={{RULER: What's the Real Context Size of Your Long-Context Language Models?}}, 
      author={Cheng-Ping Hsieh and Simeng Sun and Samuel Kriman and Shantanu Acharya and Dima Rekesh and Fei Jia and Yang Zhang and Boris Ginsburg},
      year={2024},
      eprint={2404.06654},
      archivePrefix={arXiv},
      primaryClass={cs.CL},
      url={https://arxiv.org/abs/2404.06654}, 
}

@misc{lee2024,
      title={{Can Long-Context Language Models Subsume Retrieval, RAG, SQL, and More?}}, 
      author={Jinhyuk Lee and Anthony Chen and Zhuyun Dai and Dheeru Dua and Devendra Singh Sachan and Michael Boratko and Yi Luan and Sébastien M. R. Arnold and Vincent Perot and Siddharth Dalmia and Hexiang Hu and Xudong Lin and Panupong Pasupat and Aida Amini and Jeremy R. Cole and Sebastian Riedel and Iftekhar Naim and Ming-Wei Chang and Kelvin Guu},
      year={2024},
      eprint={2406.13121},
      archivePrefix={arXiv},
      primaryClass={cs.CL},
      url={https://arxiv.org/abs/2406.13121}, 
}

@article{koblischke2025,
      title={{Gravity-Bench-v1: A Benchmark on Gravitational Physics Discovery for Agents}}, 
      author={Nolan Koblischke and Hyunseok Jang and Kristen Menou and Mohamad Ali-Dib},
      year={2025},
      journal={Proceedings of the 42nd International Conference on Machine Learning (ICML)},
      url={https://arxiv.org/abs/2501.18411}, 
}

@article{ji2021,
    author = {Ji, Yanrong and Zhou, Zhihan and Liu, Han and Davuluri, Ramana V},
    title = {{DNABERT: pre-trained Bidirectional Encoder Representations from Transformers model for DNA-language in genome}},
    journal = {Bioinformatics},
    volume = {37},
    number = {15},
    pages = {2112—2120},
    year = {2021},
    issn = {1367—4803},
    doi = {10.1093/bioinformatics/btab083},
    url = {https://doi.org/10.1093/bioinformatics/btab083},
    eprint = {https://academic.oup.com/bioinformatics/article-pdf/37/15/2112/57195892/btab083.pdf}
}

@article{zhou2024,
      title={{DNABERT-2: Efficient Foundation Model and Benchmark For Multi-Species Genomes}}, 
      author={Zhihan Zhou and Yanrong Ji and Weijian Li and Pratik Dutta and Ramana Davuluri and Han Liu},
      year={2024},
      journal={Proceedings of the 12th International Conference on Learning Representations (ICLR)},
      url={https://arxiv.org/abs/2306.15006}, 
}

@article{
    brixi2026,
    author = {Brixi, Garyk and Durrant, Matthew G. and Ku, Jerome and Naghipourfar, Mohsen and Poli, Michael and Sun, Gwanggyu and Brockman, Greg and Chang, Daniel and Fanton, Alison and Gonzalez, Gabriel A. and King, Samuel H. and Li, David B. and Merchant, Aditi T. and Nguyen, Eric and Ricci-Tam, Chiara and Romero, David W. and Schmok, Jonathan C. and Taghibakhshi, Ali and Vorontsov, Anton and Yang, Brandon and Deng, Myra and Gorton, Liv and Nguyen, Nam and Wang, Nicholas K. and Pearce, Michael T. and Simon, Elana and Adams, Etowah and Amador, Zachary J. and Ashley, Euan A. and Baccus, Stephen A. and Dai, Haoyu and Dillmann, Steven and Ermon, Stefano and Guo, Daniel and Herschl, Michael H. and Ilango, Rajesh and Janik, Ken and Lu, Amy X. and Mehta, Reshma and Mofrad, Mohammad R. K. and Ng, Madelena Y. and Pannu, Jaspreet and Ré, Christopher and St. John, John and Sullivan, Jeremy and Tey, Joseph and Viggiano, Ben and Zhu, Kevin and Zynda, Greg and Balsam, Daniel and Collison, Patrick and Costa, Anthony B. and Hernandez-Boussard, Tina and Ho, Eric and Liu, Ming-Yu and McGrath, Thomas and Powell, Kimberly and Pinglay, Sudarshan and Burke, Dave P. and Goodarzi, Hani and Hsu, Patrick D. and Hie, Brian L.},
    title = {Genome modelling and design across all domains of life with Evo 2},
    journal = {Nature},
    volume = {652},
    number = {8112},
    pages = {1349—1361},
    year = {2026},
    doi = {10.1038/s41586-026-10176-5},
    URL = {https://doi.org/10.1038/s41586-026-10176-5}
}

@article{
    hayes2025,
    author = {Thomas Hayes  and Roshan Rao  and Halil Akin  and Nicholas J. Sofroniew  and Deniz Oktay  and Zeming Lin  and Robert Verkuil  and Vincent Q. Tran  and Jonathan Deaton  and Marius Wiggert  and Rohil Badkundri  and Irhum Shafkat  and Jun Gong  and Alexander Derry  and Raul S. Molina  and Neil Thomas  and Yousuf A. Khan  and Chetan Mishra  and Carolyn Kim  and Liam J. Bartie  and Matthew Nemeth  and Patrick D. Hsu  and Tom Sercu  and Salvatore Candido  and Alexander Rives },
    title = {{Simulating 500 million years of evolution with a language model}},
    journal = {Science},
    volume = {387},
    number = {6736},
    pages = {850—858},
    year = {2025},
    doi = {10.1126/science.ads0018},
    URL = {https://www.science.org/doi/abs/10.1126/science.ads0018},
    eprint = {https://www.science.org/doi/pdf/10.1126/science.ads0018}
}

@misc{vdo2018,
      title={{Neural Discrete Representation Learning}}, 
      author={Aaron van den Oord and Oriol Vinyals and Koray Kavukcuoglu},
      year={2018},
      eprint={1711.00937},
      archivePrefix={arXiv},
      primaryClass={cs.LG},
      url={https://arxiv.org/abs/1711.00937}, 
}

@article{doniach1970,
doi = {10.1088/0022-3719/3/2/010},
url = {https://doi.org/10.1088/0022-3719/3/2/010},
year = {1970},
month = {feb},
publisher = {},
volume = {3},
number = {2},
pages = {285},
author = {S Doniach and M Sunjic},
title = {Many-electron singularity in X-ray photoemission and X-ray line spectra from metals},
journal = {Journal of Physics C: Solid State Physics}
}

@article{norman1998,
  title = {Phenomenology of the low-energy spectral function in high-${T}_{c}$ superconductors},
  author = {Norman, M. R. and Randeria, M. and Ding, H. and Campuzano, J. C.},
  journal = {Phys. Rev. B},
  volume = {57},
  issue = {18},
  pages = {R11093(R)--R11096(R)},
  numpages = {0},
  year = {1998},
  month = {May},
  publisher = {American Physical Society},
  doi = {10.1103/PhysRevB.57.R11093},
  url = {https://link.aps.org/doi/10.1103/PhysRevB.57.R11093}
}

%%%%%%%%%%%%%%%%%%%%%%%%%%%%%%%%%%%%%%%%%%%%%%%%%%%%%%%%%%%%

\clearpage
\newpage
\section*{Appendix A\quad Previous Benchmarks and Further Work} \label{appendixA}

Developing benchmarks to test the manifold capabilities of AI has been an area of serious inquiry since its inception \citep{turing50}. Early benchmarks drew upon specially collected, publicly available datasets, and tackled specific tasks: MNIST \citep{deng2012} for visual character recognition, ImageNet \citep{deng2009} for object classification, SQuAD \citep{rajpurkar2016} for reading comprehension, and BLEU \citep{papineni2002} for machine translation. A trend towards increasing comprehensiveness is discerned, notably in GLUE \citep{wang2018} and SuperGLUE \citep{wang2019} benchmarks for natural language processing (NLP), as well as breadth, such as in the UCI Machine Learning Repository \citep{kelly2023}, MMLU \citep{hendrycks2021}, and GSM8K \citep{cobbe2021}.

The recent development of LLMs has spawned increasingly sophisticated benchmarks. Chatbot Arena \citep{chiang2024} pioneered the crowd–sourcing of chatbot evaluation to the public, which was extended in SciArena \citep{zhao2025} to scientific literature tasks evaluated by the scientific community. The rigor of benchmarks has also improved considerably, evidenced by the increased attention paid to the flaws of earlier generations of benchmarks. A few examples in NLP suffice: Holistic Evaluation of Language Models (HELM) \citep{liang2023} recognized diverse preferences of evaluators in vastly broadening the scope of metrics applied, AlpacaEval \citep{dubois2025} tackled the bias of auto–evaluators in favor of longer answers, and LiveBench \citep{white2025} minimized the effect of test set contamination by continually refreshing its corpus of tasks. The scope of tasks evaluated has also evolved, particularly towards real–world tasks in WildBench \citep{lin2024} and professional coding in SWE–bench \citep{jimenez2024}. In parallel, the level of domain expertise tested has grown markedly, such as GPQA \citep{rein2023} and Humanity's Last Exam \citep{phan2025} covering graduate–level tasks across diverse fields and FrontierMath \citep{glazer2024} targeting expert–level mathematical problems. The past year has seen the emergence of specialized, GPQA–style benchmarks (that is, in the format of graduate school–level examinations) written by domain experts, including the field of physics \citep{chung2025} and condensed matter in particular \citep{wang2025}. 

We highlight relevant \textbf{benchmarks outside of science} evaluating various capabilities: inductive reasoning in InductionBench \citep{hua2025} and ARC–AGI \citep{chollet2025}, information extraction from complex data in visual reasoning benchmarks, and noise robustness in NoiseQA \citep{ravichander2021}. In particular, long context processing was assessed in Long Range Arena \citep{tay2020}, and at a more sophisticated level in Michelangelo's Latent Structure Query framework \citep{vodrahalli2024}, OpenAI's adapted Multi–Round Co–reference Resolution (MRCR) \citep{openai2025}, GSM–$\infty$ \citep{zhou2025}, and MMReD \citep{kurkin2025}.

\textbf{We anticipate that several developments are necessary before the full potential of agents is harnessed in condensed matter.} \textit{Firstly}, agents should become fluent in the `domain language' of any field to be able to represent a system in its full complexity and be poised to make profound inferences, compared to agents with a merely superficial grasp. Having found that code generation by agents is still insufficient, we anticipate fine–tuning models through RL as a next step, or giving agents access to train other (possibly simpler) models \citep{ying2022}. In the future, we may need to construct domain–specific foundation models, or even find that transformer–based LLMs themselves are insufficient, calling for new architectures such as world models \citep{lecun2022}. The importance of choosing the right representation (such as difference between structure– and sequence–based models in protein folding) cannot be ignored. For example, vision language models may handle data in a compressed way as opposed to enumerating all intensities numerically and in all, but quantitative reasoning is often lost.

\textit{Secondly}, agents must be able to reason through different layers of representations. Due to the ubiquity of \textbf{emergence} in condensed matter \hyperref[appendixD]{(Appendix D)}, the microscopic origin at which theories work is typically \textit{many (qualitatively) different layers of reasoning removed} from measurable properties, the latter of which are the starting point of data analysis. The path from experiment to theory is therefore highly serpentine and at times impenetrable. (ARPES and RIXS are special in that the number of these `layers' involved is fewer than most other experimental techniques.) The scientific process, especially in condensed matter, consists not of a \textit{single} inverse problem (Tier I and II questions in CME–SpectrumBench), but of a \textit{string} of inverse problems, in which the description of the system is couched in a different representation along each step. The ability to start from raw data and reason through to the level of microscopic theory, may be a necessary precondition for a model to tackle the problem of emergence. Until that point is reached, human intervention is expected at every step to guide models through these disparate layers of reasoning.

\textit{Lastly}, an agent has to be capable of generalizing and reasoning \textit{across} the results of various measurements to piece together a physical picture that transcends the mere specifics of individual experiments — today, this is achieved through the formation of a scientific consensus across a broad array of human experts \citep{polanyi1962}. This is because an \textit{irreversible loss of information} occurs in data collection, as the complexities of microscopic physics are collapsed into the highly constrained feature space that each measurement affords. The parable of the blind men and the elephant is apt. From a \textit{single} experiment alone, one cannot uniquely trace any phenomenon back to a single underlying microscopic origin, therefore a holistic description of a system or phenomenon (such as a superconductor) is necessarily pieced together from \textit{disparate} experiments; even this collected information provides but a fragmentary understanding. We hope that agents in the medium–term future, possibly acting in synergy with human guidance, may supplement this process by performing a variety of targeted measurements to compare with, and constrain possible physical models. In superconductivity, for example, structural probes (such as X–ray diffraction) and electronic probes (such as ARPES) may be compared with theoretical predictions arising from microscopic mechanisms (phonons, spin fluctuations, excitonic pairing). To maintain rigor in the scientific process, and achieve convergence between theory and experiment at each step, critical for preserving a coherent overall understanding, each of these quantities must be benchmarked and calibrated with experiments. \\

\section*{Appendix B\quad The Scientific Process and Condensed Matter} \label{appendixB}

The scientific process is an iterative cycle of interrelated tasks: \textit{hypothesis generation, experiment execution, data analysis, and communication}. Despite efforts to automate the full research pipeline \citep{jansen2025, lu2024, yamada2025}, no LLM to date can duplicate the full pipeline with the same degree of accuracy and fine–grained detail as human research teams. Consequently, \textbf{scientific benchmarks} test either the whole process itself via well–designed toy models of the pipeline \citep{koblischke2025}, or specific tasks in that cycle. Regular scientific use of LLMs has shifted beyond mere brainstorming/writing aids \citep{liang2025}, and recent benchmarks have sprung up measuring the competence of models elsewhere in the scientific process, particularly by crystallizing them as code generation \citep{chen2024} and workflow derivation problems \citep{majumder2024}. The question of whether data analysis can be tackled had been treated by \citet{liu2024a} on smaller, non–scientific datasets.

\textbf{CME–SpectrumBench} is targeted at the weakest link in the scientific process: it asks whether models can be \textit{trusted} on data analysis under minimal prompt information and human intervention (no elicitation), while maintaining full standards of rigor, accountability, and interpretability. The role of an agent or human in scientific discovery lies in the act of `controlled rebellion' \citep{polanyi1962}: the \textit{originality} (degree of surprise) of their discovery is held in tension against high standards of \textit{proof} enforced by the conservative attitude of science towards modifying consensus — standards judged and upheld by other human scientists. Any agent seeking to supplement, let alone supplant, the work of a human in science must \textit{first} match the same high bar, or find its results disregarded by the scientific community.

CME–SpectrumBench is focused on condensed matter. \textbf{Condensed matter} may be compared to a river or strait that is \textit{very long} (it has a diversity of problems) and \textit{wide} (theoretical models on one bank, experimental observations on the other). When bridging theory and experiment, one begins with winnowing down the extreme complexity of a physical system by identifying relevant microscopic degrees of freedom. A hypothesized model Hamiltonian can then be written down and solved with mathematical tools at hand. One can then calculate physical quantities and compare them with experimental results. The process iterates until a satisfactory theory (Hamiltonian) is reached that consistently captures the essence of a physical system.

The `bridge' may be broadly divided into two halves, like the two towers of the Golden Gate Bridge. Recent benchmarks, notably \citep{pan2025}, \citep{wang2025a}, and \citep{wang2025}, have tackled the \textbf{\textit{theoretical}} side, where difficulties faced by researchers are mathematical and computational. In turn, \textit{we are the first to deal with the other, \textbf{experimental} side}, where research problems stem from sample preparation, instrumental difficulties, and analysis of data. We focus on the most powerful experimental tools in the arsenal of condensed matter — the spectroscopic probes of ARPES and RIXS, the narrowest spot of the condensed matter `river', thus the best place to build a `bridge'. Models performing well on CME–SpectrumBench can automate analytical work for this field, and results may generalize to other spectral data analysis tasks in physics and chemistry. We note three strengths of our benchmark.\begin{enumerate}
    \item Of the limited number of techniques available, two spectral probes, ARPES and RIXS, stand out for their \textbf{rich, extended feature space and more direct linkage to Hamiltonians}, relying less on theoretical scaffolding for their interpretation. Domain–specific scientific foundation models emphasize the importance of \textbf{learning the `domain language' in which data is represented} \citep{zhang2025}: genomes in biology \citep{nguyen2024}, molecules in chemistry \citep{chithrananda2020, kim2021}. Electronic structure, measured by ARPES, is a key `domain language' (or \textit{`genomic code'}) of condensed matter, telling us how electrons behave in materials, thus explaining their properties. A similar role is played by RIXS for elementary excitations. Understanding electronic structure and excitations is necessary for learning an informative representation of material systems, and crucial \citep{goyal2025} for constructing foundation models for the critical field of novel materials — the reason why condensed matter is important to technological policy, such as the Genesis Mission \citep{genesis2025}. Because \textit{grokking} spectral data is the primary way scientists currently use to reveal electronic structure \citep{yang2018}, the ability to process spectral data is essential to leveraging the capabilities of frontier models to contribute decisively to this sector of technological advancement.
    \item The core challenges of ARPES and RIXS—finding patterns in noisy, multi–dimensional datasets—recur across many subfields in physics and is a \textbf{good proxy for other experimental techniques}, particularly where spectra and images are analyzed. Spectral datasets are large enough to challenge context window limits, while not being excessively high–dimensional that rudimentary tasks exceed LLM limits, as in  calorimetric data in high–energy physics \citep{baldi2016}.
    \item The use of LLMs for research tasks in condensed matter is at an early stage compared to other scientific fields such as protein modeling. By identifying which problems in our benchmark have occurred elsewhere (especially involving long-context reasoning), \textbf{we may repurpose solutions originally designed in other contexts for CME}. \\
\end{enumerate}

\section*{Appendix C\quad Extensions to Other Experimental Techniques} \label{appendixC}

This appendix presents general guidelines for other experimentalists wishing to evaluate the capabilities of agents on analyzing their own data, in the same manner as CME–SpectrumBench.

\paragraph{Format of prompts} In adapting CME–SpectrumBench to various experimental techniques, prompts should contain only a modicum of information necessary for agents to understand the nature of the data they are presented, without being assisted by additional information. Researchers with expertise in different fields are disposed to phrase questions very differently, and it is crucial to ensure that in extending the scope of CME–SpectrumBench, the test remains a pure evaluation of \textit{reasoning} abilities of agents themselves, rather than partly be a test of \textit{elicitation} capabilities by the writers of prompts. The format of questions should follow that set out in \hyperref[section2.2]{Section 2.2}. Several guidelines apply: \begin{itemize}
    \item Questions should be written in \texttt{.txt} and structured in the form \texttt{prompt + content}, with all data contained within \texttt{content}.
    \item In lieu of a detailed explanation in the \texttt{prompt} on how data should be analyzed, agents should deduce this method through few–shot prompting: multiple examples are given (in \texttt{content}) with their answers stated (in \texttt{prompt}), before the actual dataset to be analyzed is stated (obviously without the answer).
    \item The expected form and length of the answer should also be stated in the prompt: whether it is a single numerical value or an array of values; for the latter, the length of the array should also be stated. This is necessary for analysis to be automated, while agents with structured responses are not unfairly advantaged over agents without.
    \item Data should be either one– or two–dimensional, with axis values, units, and labels stated. Two--dimensional data should presented as a table; the explanation in \hyperref[section2.2]{Section 2.2} suffices.
    \item Precision of numerical values in the data may be left to the discretion of the evaluator. In our ARPES case (which may generalize to other methods with high–resolution, two–dimensional data), the context window afforded by an agent may be easily exceeded, and we found it necessary to limit the precision of our data in the manner mentioned in \hyperref[section2.2]{Section 2.2}.
\end{itemize}

\paragraph{Tiering of questions} We recognize that different experimental probes involve vastly different issues when it comes to the question of interpretability. ARPES and RIXS are unusual in that variations in intensity in their data may be straightforwardly ascribed to simple excitations. That is because it has a comparatively \textit{large feature space} (energy and several momentum axes) which are collapsed into a smaller set of axes in many other measurements: for example, many other spectroscopies (such as Raman, or Angle–Integrated Photoemission) are not resolved in momentum space, and retain only the energy axis (the `spectrum', hence `spectroscopy'). Furthermore, many transport measurements (such as resistivity and heat conductivity) lack even energy resolution; instead, data consists of the variation of some quantity with another (such as temperature).

Nevertheless, a tiering of questions by difficulty (\hyperref[comparingtiers]{Comparing tiers} in Section 4.2) may be generalized to other measurements. Departing from the set of difficulty tiers stated in \hyperref[section2.3]{Section 2.3}, examples are given for other experimental techniques, with prior ARPES and RIXS examples (\hyperref[appendixF]{Appendices F} and \hyperref[appendixG]{G}) italicized.\begin{itemize}
    \item Tier I — extraction of \textbf{a single quantity} departing to a limited extent from NIAH type questions. Examples: \textit{Fermi level, band bottom energy, Dirac cone energy, superconducting gap size, phonon frequencies, RIXS category F}, positions of peaks, minima, and discontinuities in 1D arrays such as resistivity/conductivity, specific heat, magnetic susceptibility, quantum oscillations, and most spectroscopies presenting only energy resolution.
    \item Tier II — extraction of \textbf{an array of quantities}. Examples: \textit{tracing 1D dispersions} $\epsilon_k$ \textit{and line widths} $\Sigma''(\omega)$ \textit{from a 2D data array, RIXS category G}, variation in superconducting gap size along a 1D path in tunneling measurements, edge states in microwave impedance microscopy.
    \item Tier III — single quantities that are indirectly determined; that is, \textbf{quantities calculated after extracting} such arrays as those of Tier II, or parsing arrays directly stated in the data. Examples: \textit{Fermi velocities, doping levels}, scaling laws in transport measurements and what processes they may be ascribed to (sources of scattering), scattering wavevectors obtained by Fourier transforming quasiparticle interference data, matching Raman frequencies obtained from the data with their possible origins (such as specific phonons).
\end{itemize}

\paragraph{Experimental complications} It is recommended that data be simulated, with expected experimental complications inserted. This makes sure that (1) a predetermined ground truth exists (and is not restrained by the potential inaccuracy and slow speed of human data analysis), and that (2) the intensity of complicating factors may be quantified. In particular, we watch out for: \begin{itemize}
    \item \textit{Noise levels}. In our benchmark, the intensity of noise is measured as a fraction of the \textit{peak signal} intensity, rather than alternatives such as average intensity of the entire dataset, which depends on such other factors as background level and axis ranges. A similar prescription is recommended. (Our only deviation from this is in Fermi level extraction from a featureless background, for which the background itself constitutes most of the `signal').
    \item \textit{Convolution}. It is evident that broader convolution increases the difficulty for a signal to be extracted. While we have not conducted a detailed study on this dependence (as we had for noise), we have found it sufficient to fix convolution at a level expected in an actual experiment, and measure deviation with respect to the half–width half maximum (HWHM) of the convolution \hyperref[section3.2]{(Section 3.2)}. \\
\end{itemize}

\section*{Appendix D\quad Emergence in Condensed Matter} \label{appendixD}

The core idea that emergent behavior is central to the description of large systems is a point of commonality between the artificial intelligence and condensed matter communities. This is the belief that knowledge of the properties or laws that govern \textit{smaller components} (such as atoms in physics, or artificial neurons in machine learning) are at best ancillary to, and at worst exceedingly insufficient for predicting the behavior of \textit{large systems} composed of them (by the same analogy—a material, or a neural network). It stands in (partial) opposition to the reductionist impulse that had historically prevailed over much of physics, and is embodied in AI thinking by such approaches as mechanistic interpretability. We provide several examples of emergence for readers in the AI community.
\begin{itemize}
    \item \textbf{Criticality and scaling laws.} As a system approaches a phase transition (such as boiling water), fluctuations become increasingly large (size of bubbles), rendering most microscopic details irrelevant to its behavior. This gives rise to scaling laws that relate thermodynamic quantities and depend on very few properties. Vastly different systems that happen to share these properties (more precisely, dimension and symmetry) then develop the same scaling laws; this is known as \textbf{universality} \citep{stanley1971}. An example is the onset of ferromagnetism and boiling water (critical opalescence), both of which are classed into the Ising model. The idea that scaling laws emerge in large systems is also central to the AI mythos, epitomized by the work of \citet{kaplan2020}, which show how performance (loss) scaled with some measure of size (such as parameters, tokens, or compute) — a universal phenomenon, because of its independence of model architecture. A key difference exists between the two communities. Scaling laws in AI often revolve around some aspect of system size, as this approach satisfies the need to improve performance far more than varying architectural details (for which similar laws exist). On the other hand, physicists, whose attention is not focused on any single overarching goal in scaling, consider a wider range of laws.
    \item \textbf{Superconductivity.} This is an inherently quantum mechanical phenomenon and an example of a collective instability. When conditions arise in which electrons experience a net attraction rather than repulsion, and pair up (Cooper pairing), a lower-energy ground state is favored. This ground state is the superconducting state; it manifests in ARPES spectra as the opening of an energy gap around the Fermi level (see question \texttt{B6} in \hyperref[appendixF]{Appendices F} and \hyperref[appendixG]{G}). Although the mechanism for Cooper pairing is microscopic, the superconducting state exhibits such macroscopic properties as zero resistance, expulsion of magnetic fields, and macroscopic quantum coherence.
    \item \textbf{Non–Fermi liquids.} In a Fermi liquid (such as many metals), interactions between electrons are weak enough that the qualitative behavior of the whole system departs little from a hypothetical system that lacks these interactions (the independent electron approximation), albeit with some numerics altered (renormalization). This is because a one–to–one mapping between states of these two systems (quasiparticles and independent electrons, respectively) is retained; a superficial analogy may be made with the process of tokenization in LLMs. In a \textit{non}–Fermi liquid, however, the mapping breaks down, and the system produces \textit{qualitatively different} behavior. This is partly because the fundamental constituents of the system have ceased to be quasiparticles; what they are is presently unknown. In a sense this is a problem of \textit{representation}. Those in the AI community grappling with non–interpretable features in large models are dealing with a comparable matter. \\
\end{itemize}

\section*{Appendix E\quad Scoring System} \label{appendixE}

It was stated in \hyperref[section3.2]{Section 3.2} that a Lorentzian (otherwise known as Cauchy) distribution was used,\[\textrm{Score} = \gamma^2\boldsymbol{\cdot}\big[(x - x_0)^2 + \gamma^2\big]^{-1}\]with choices of the half–width half–maximum $\gamma$ depending on the quantity being measured. We recognize that more sophisticated measures of deviation (or even distance) may be in principle be used, and that there is no \textit{a priori} reason why a Lorentzian should be selected over, say, a Gaussian; its `long tail' merely allows us to pick up responses deviating further from the ground truth. There is no mathematically rigorous reason why the Lorentzian distribution should be favored over others, but two general principles apply:\begin{itemize}
    \item Completely correct answers should receive a score of 1 and completely incorrect answers a score of 0; that is, there is a finite bound of scores, and this bound is normalized to unity. Our intention is to reward answers which come close to the ground truth without unnecessarily penalizing those which stray far from it. An answer that is \textit{rather} wrong and an answer that is \textit{very} wrong would \textit{both} attain similar, near–zero scores. This rules out the use of loss or reward functions which are not bounded, as they discriminate between increasingly incorrect answers in their task to guide models towards optimal solutions. (Inverting this reason, future extensions of CME–SpectrumBench tailored to performing reinforcement learning on open–weight models may replace bounded scores with unbounded loss functions.)
    \item The score monotonically decreases with increasing deviation. A different measure abiding by this requirement could have been chosen, such as a Gaussian (whose choice could have been supported by the central limit theorem), but this choice only approximately amounts to a redistribution of weighting between answers of different accuracy, and a similar change could be effected by just changing the value of $\gamma$. A Lorentzian was ultimately chosen for two reasons: (1) it has a longer tail compared to the Gaussian, which is equivalent to more relaxed scoring, and (2) it is the Lorentzian and not Gaussian distribution that naturally appears in spectral line shapes such as those seen in ARPES and RIXS, where half–width half–maxima $\gamma$ do attain physical meaning. Scientists extending this benchmark to other fields may wish to use the Gaussian if they see fit, especially if its use is found to be more justified within their domain. \\
\end{itemize}

\section*{Appendix F\quad Question Categories, Example Scripts, and Reproducibility} \label{appendixF}

\subsection*{Question categories}

The 38 categories of questions mentioned in \hyperref[section2.3]{Section 2.3} are grouped into seven domains (A--G) according to the physical quantity of interest. These are shown in Figs.~\ref{fig:a}--\ref{fig:g} with accompanying explanations. Note that all spectra are shown at the highest resolution setting (at the upper end of the scale in Fig.~\ref{fig:ablations}(a)) and \textit{without} noise, or \textit{minimal} convolution. Full scores for all models tested, across all 190 questions (multiplexed by five levels of noise), are shown in Figs.~\ref{fig:direct} and \ref{fig:advanced}, with explanatory comments. These seven domains of scientific significance are listed, with five ARPES (\texttt{A-E}) and two RIXS (\texttt{F-G}) domains: \begin{itemize}
    \item \textbf{A.} Extraction of Fermi level (\texttt{A1}).
    \item \textbf{B.} Extraction of dispersions $\epsilon(k)$ or information thereof. Questions requiring array responses include linear (\texttt{B1}) and curved/quadratic (\texttt{B2}) dispersions; linear dispersions with superstructure (\texttt{B3}) and displaying band bottoms (\texttt{B4}). Questions requiring single numerical answers are the Fermi velocities ($v_{\textrm{F}}$) of linear (\texttt{B1\_vF}), curved (\texttt{B2\_vF}), and superstructural (\texttt{B3\_vF}) dispersions, as well as the energies of band bottoms (\texttt{B4\_bbE}), Dirac cones (\texttt{B5}), and superconducting gaps (\texttt{B6}).
    \item \textbf{C.} Extraction of line widths $\Sigma''(\omega)$ as arrays, whose variation with energy is an indication of interaction processes in a material: impurity scattering (\texttt{C1}), marginal Fermi liquid/MFL (\texttt{C2}), Fermi liquid/FL (\texttt{C3}), MFL and a single phonon (\texttt{C4}), FL with a single phonon (\texttt{C5}).
    \item \textbf{D.} Retrieval of phonon energy, whose presence is revealed by a kink in the dispersion $\epsilon(k)$ at the phonon energy and increase in line width $\Sigma''(\omega)$ approaching and passing below the phonon energy. This includes a single phonon (\texttt{D1}) at five levels of coupling strength $\lambda$ (showing up as increasing salience of the aforementioned phonon features), as well as two (\texttt{D2}) and three (\texttt{D3}) phonons at a fixed, intermediate coupling strength.
    \item \textbf{E.} Extraction of doping level for a single--band cuprate (\texttt{E1}), two--band cuprate (\texttt{E2}), strontium ruthenate (\texttt{E3}), and three--band nickelate (\texttt{E4}).
    \item \textbf{F.} Extraction of single quantities in RIXS, such as from energy loss–incident energy maps (\texttt{F1}), and single/multiple phonon frequencies (\texttt{F2-4}).
    \item \textbf{G.} Extraction of dispersions in RIXS, such as single/multiple plasmons (\texttt{G1-3}), magnon (\texttt{G4}), and multiple orbitons which have to be distinguished (\texttt{G5-7}).
\end{itemize}

The three tiers of difficulty may be mapped, as mentioned in \hyperref[section2.3]{Section 2.3}. \begin{itemize}
    \item \textbf{Tier I.} Extraction of a single quantity departing to a limited extent from NIAH questions; these are \texttt{A1}, \texttt{B4\_bbE}, \texttt{B5}, \texttt{B6}, and the $\texttt{D}$ and $\texttt{F}$ series of questions.
    \item \textbf{Tier II.} Extraction of an array of quantities, such as dispersions $\epsilon(k)$ and line widths $\Sigma''(\omega)$, including \texttt{B1}, \texttt{B2}, \texttt{B3}, \texttt{B4}, and the \texttt{C} and \texttt{G} series of questions.
    \item \textbf{Tier III.} Single quantities indirectly determined (calculated after extracting such arrays as those of Tier II). These include \texttt{B1\_vF}, \texttt{B2\_vF}, \texttt{B3\_vF}, and the \texttt{E} series of questions.
\end{itemize}

\subsection*{Example scripts}

\begin{figure}[h!]
  \centering
  \includegraphics[width=0.7\linewidth]{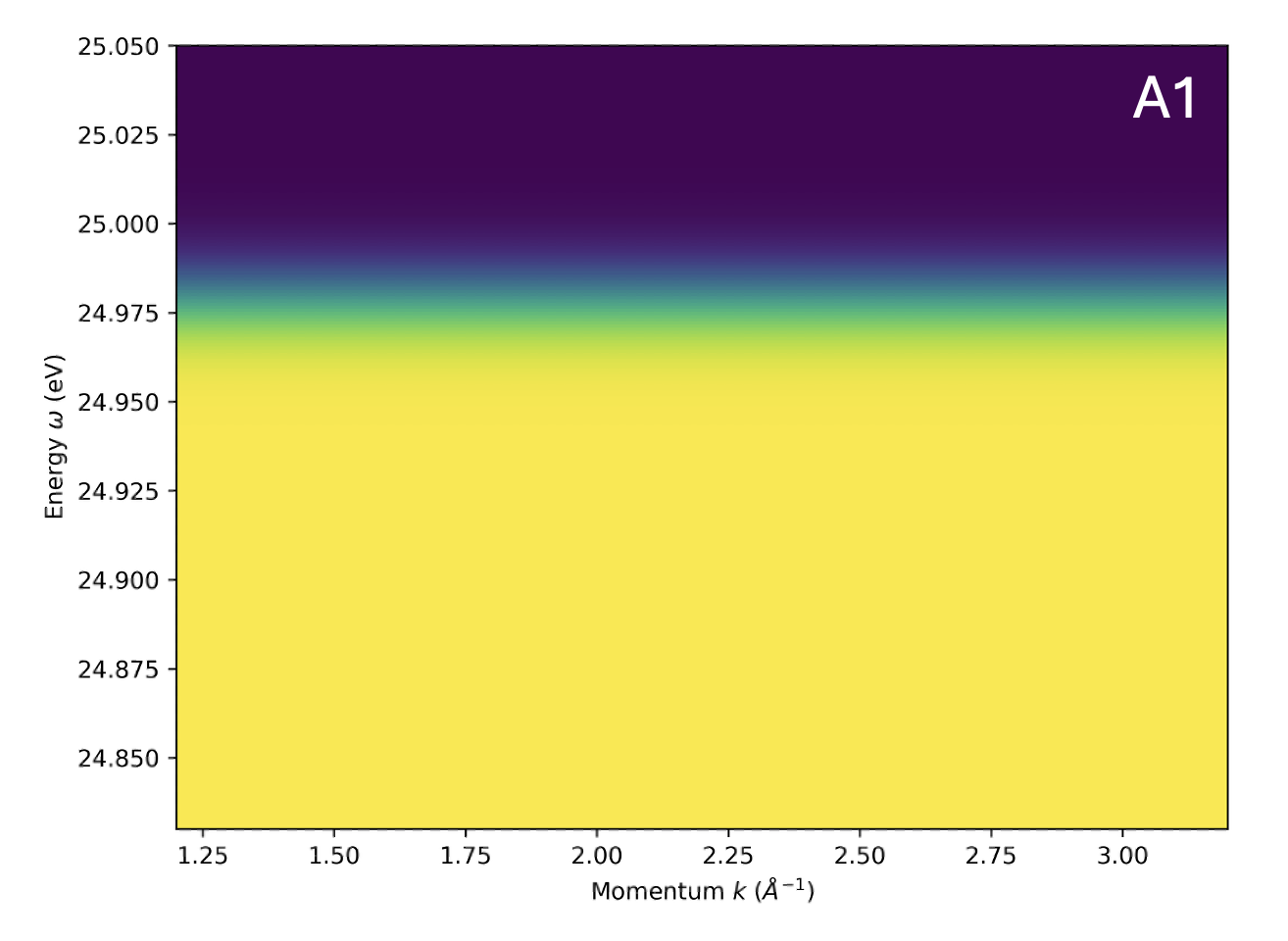}
  \caption{\textbf{Question domain A} (Fermi level extraction). Along with \texttt{F2}, this is the easiest domain, judging from scores in Figs.~\ref{fig:direct} and \ref{fig:advanced}. 
  Spectra here (\texttt{A1}) are featureless in momentum $k$ (horizontal axis), and take on a Fermi--Dirac distribution in energy $\omega$ (vertical axis),
  \\ \mbox{}\hfill$\displaystyle \textrm{Intensity}\propto \frac{1}{e^{(\omega - \mu)/k_{B} T} + 1}$\hfill\mbox{}\\
  where $\mu$ corresponds to the Fermi level, $k_{B}$ is the Boltzmann constant, and $T$ temperature.
  The Fermi level is the midpoint of spectral intensity and is visually obvious in the figure; the point of \texttt{A1} is to retrieve this energy as a single number, a task not far removed from NIAH retrieval, although the addition of noise complicates this somewhat.
  Note that spectra in question domains B--D (Figs.~\ref{fig:b}--\ref{fig:d}) have an upper cut--off at the Fermi level as little information of interest is contained above the Fermi level; this also separates the problems tested by later questions from the task of Fermi level extraction itself.
  Question domain E (Fig.~\ref{fig:e}) is arrayed along two momentum axes, cut at the Fermi level in energy $\omega$.}
  \label{fig:a}
\end{figure}

\begin{figure}[h!]
  \centering
  \includegraphics[width=\linewidth]{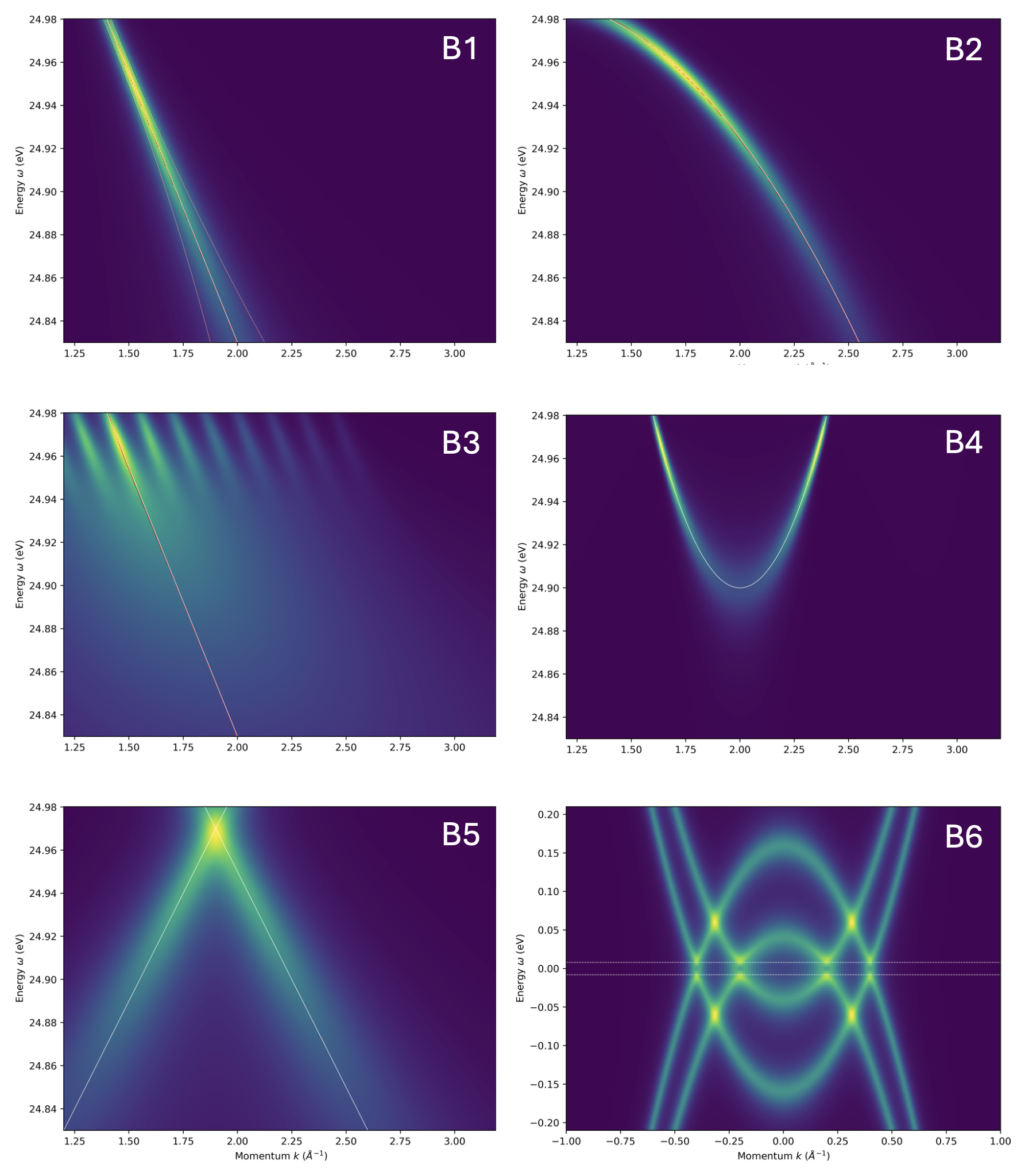}
  \caption{\textbf{Question domain B} (dispersion tracing). These cover a linear dispersion (\texttt{B1}), curved/quadratic dispersion (\texttt{B2}), linear dispersion with superstructure (\texttt{B3}), a quadratic dispersion with a band bottom (\texttt{B4}), a Dirac cone (\texttt{B5}), and superconducting gap(s) (\texttt{B6}). For \texttt{B1}--\texttt{B4}, models are prompted to trace the dispersion itself (white line) as function of momentum $k$ or energy $\omega$; these are Tier II tasks. \texttt{B4\_bbE}, \texttt{B5}, and \texttt{B6} respectively ask for the band bottom energy (bottom of parabola), Dirac cone energy (crossing point of two dispersions), and superconducting gap energy (given by half the distance between the two horizontal white lines); these are Tier I questions. Lastly, \texttt{B1\_vF}, \texttt{B2\_vF}, and \texttt{B3\_vF} ask for Fermi velocity $v_{F}$, which are the gradients $\partial\omega/\partial k$ of the dispersion at the top of the spectra; these are Tier III.}
  \label{fig:b}
\end{figure}

\begin{figure}[h!]
  \centering
  \includegraphics[width=\linewidth]{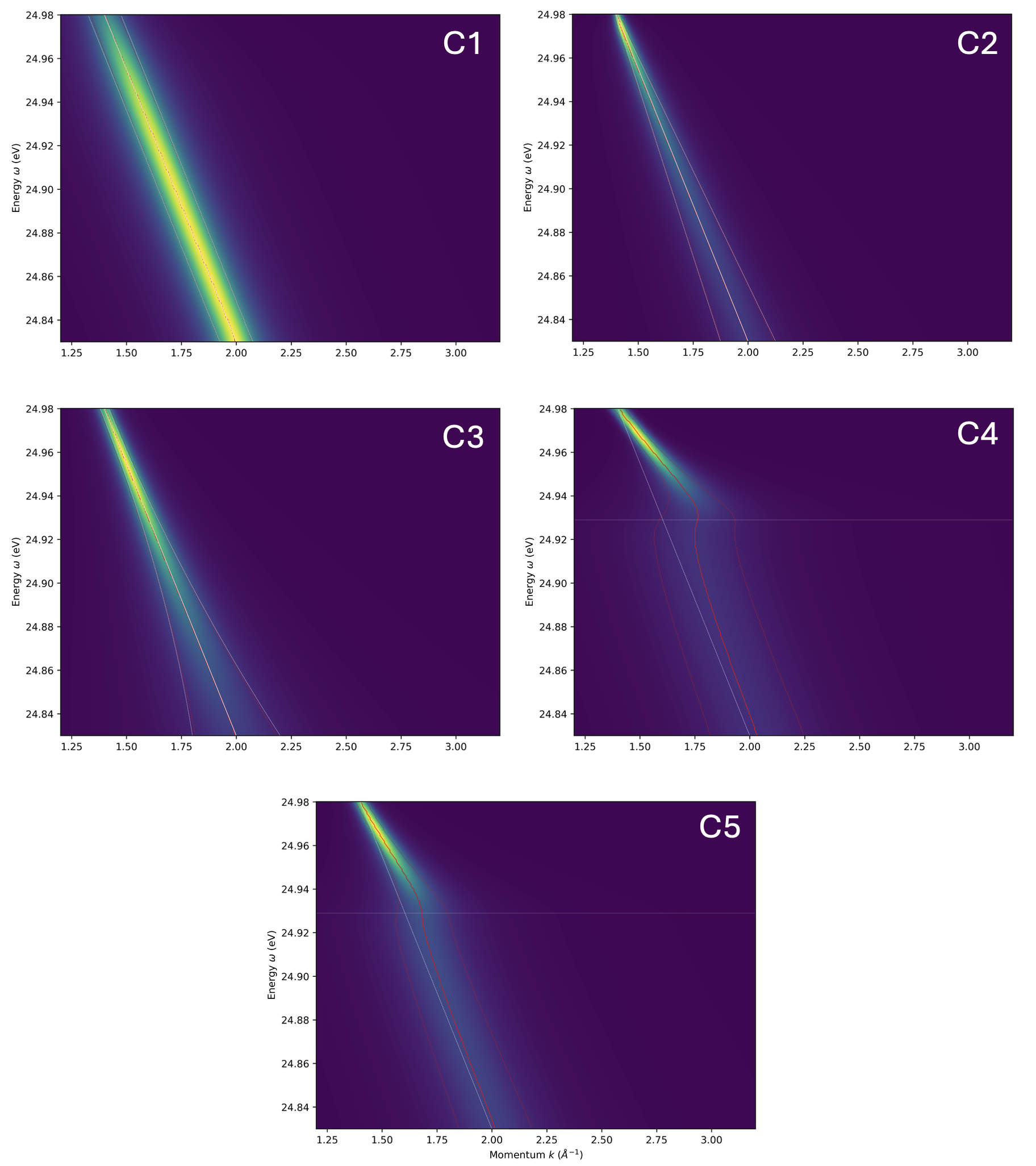}
  \caption{\textbf{Question domain C} (line width tracing). Models are prompted to reproduce half--width half--maximum (HWHM) as a function of energy $\omega$, visually shown as half the distance between the red lines. These cover cases where HWHM is constant (\texttt{C1}, impurity scattering), increases linearly away from the Fermi energy/top of spectrum (\texttt{C2}, marginal Fermi liquid), and increases quadratically (\texttt{C3}, Fermi liquid). \texttt{C4} and \texttt{C5} are variants of \texttt{C2} and \texttt{C3} with the presence of a single phonon at some phonon energy (horizontal white line). The presence of the phonon is signaled by a shift in the (red) dispersion away from the original linear (white) dispersion, an effect termed mass renormalization by physicists, and a broadening of line width below the phonon energy.}
  \label{fig:c}
\end{figure}

\begin{figure}[h!]
  \centering
  \includegraphics[width=\linewidth]{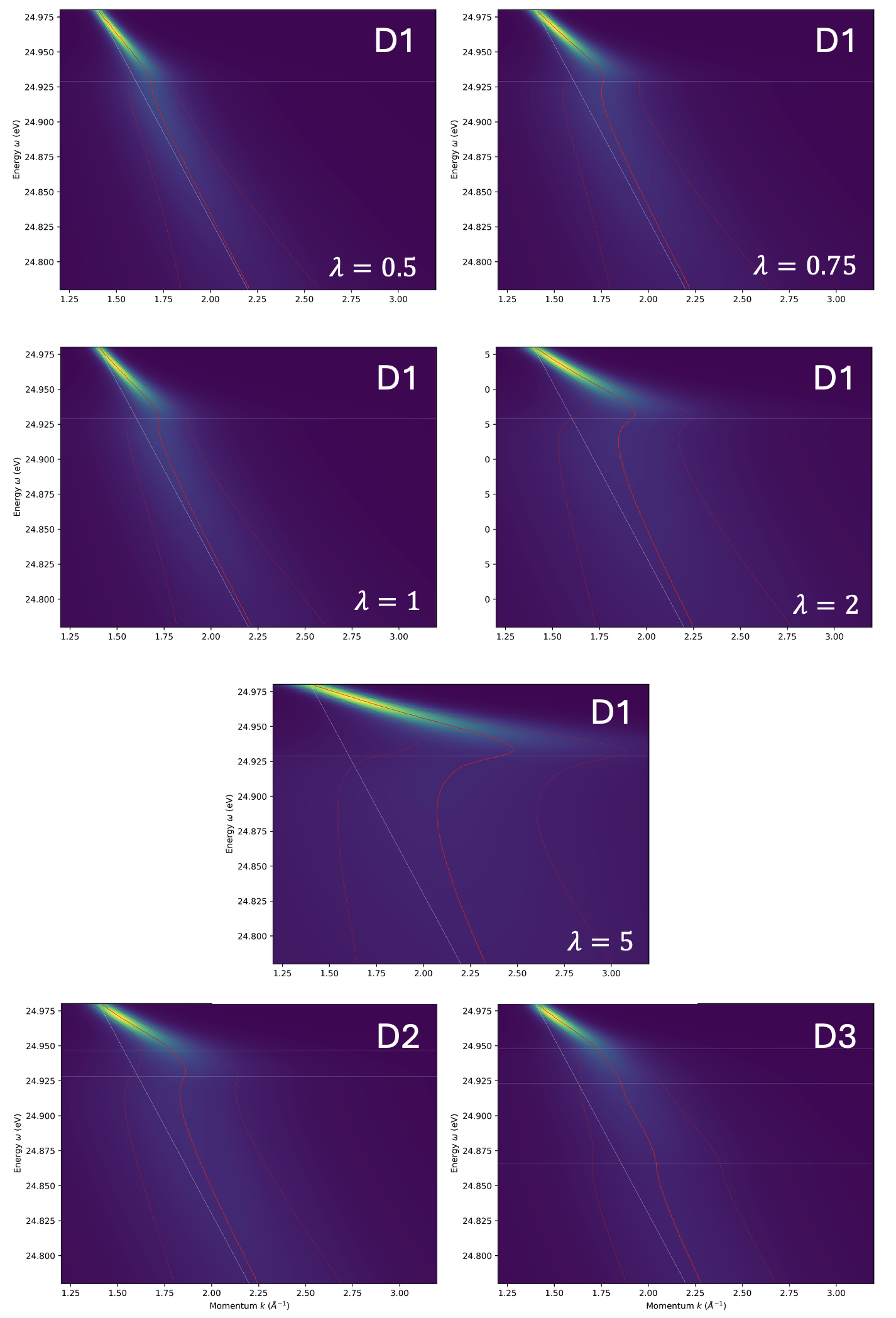}
  \caption{\textbf{Question domain D} (phonon energy determination). Horizontal white lines indicate phonon energies for a single phonon at five different coupling strengths $\lambda$ (\texttt{D1}), as well as two (\texttt{D2}) and three phonons (\texttt{D3}); models are prompted to state these energies.}
  \label{fig:d}
\end{figure}

\begin{figure}[h!]
  \centering
  \includegraphics[width=\linewidth]{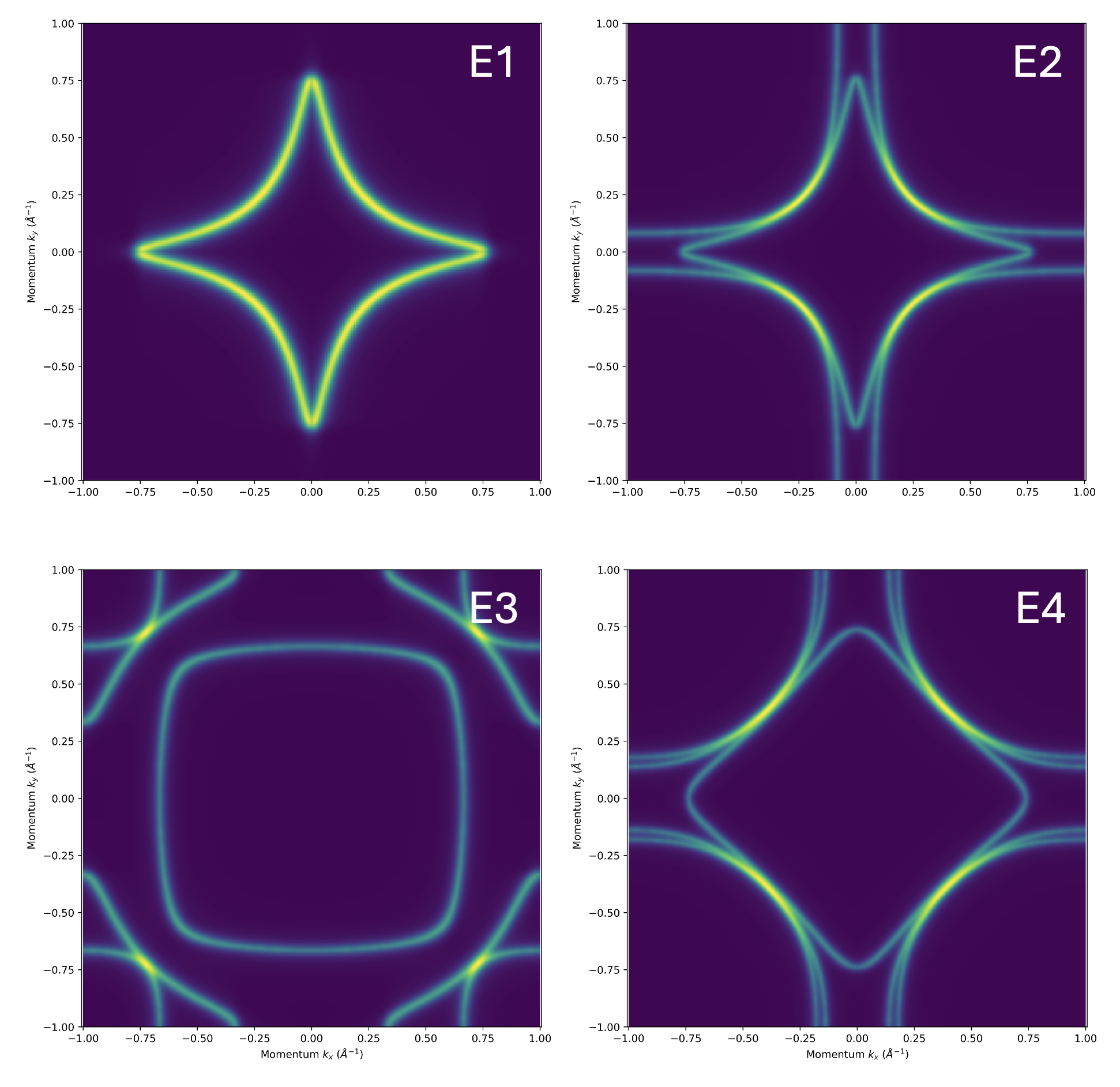}
  \caption{\textbf{Question domain E} (doping determination). Spectra shown are Fermi surface maps in two momentum axes, $k_x$ and $k_y$, over a single Brillouin zone (BZ). Fermi surfaces enclose areas whose size is linearly dependent on the doping level of a material. To keep matters internally consistent, we normalize each BZ to the same extent ($\pm 1/\textup{\AA}$ in both directions), and define \\ \mbox{}\hfill$\displaystyle \textrm{Doping} = 1 - 2\cdot\frac{\textrm{Area}}{\textrm{Area}_{(BZ)}},$\hfill\mbox{}\\ such that a Fermi surface that takes up the whole BZ has a doping of $-1$ (completely electron--doped), and none of the BZ, $+1$ (completely hole--doped). Suffice it to say that conventions and BZ sizes vary in real life, but we wish to keep things simple here.}
  \label{fig:e}
\end{figure}

\begin{figure}[h!]
  \centering
  \includegraphics[width=\linewidth]{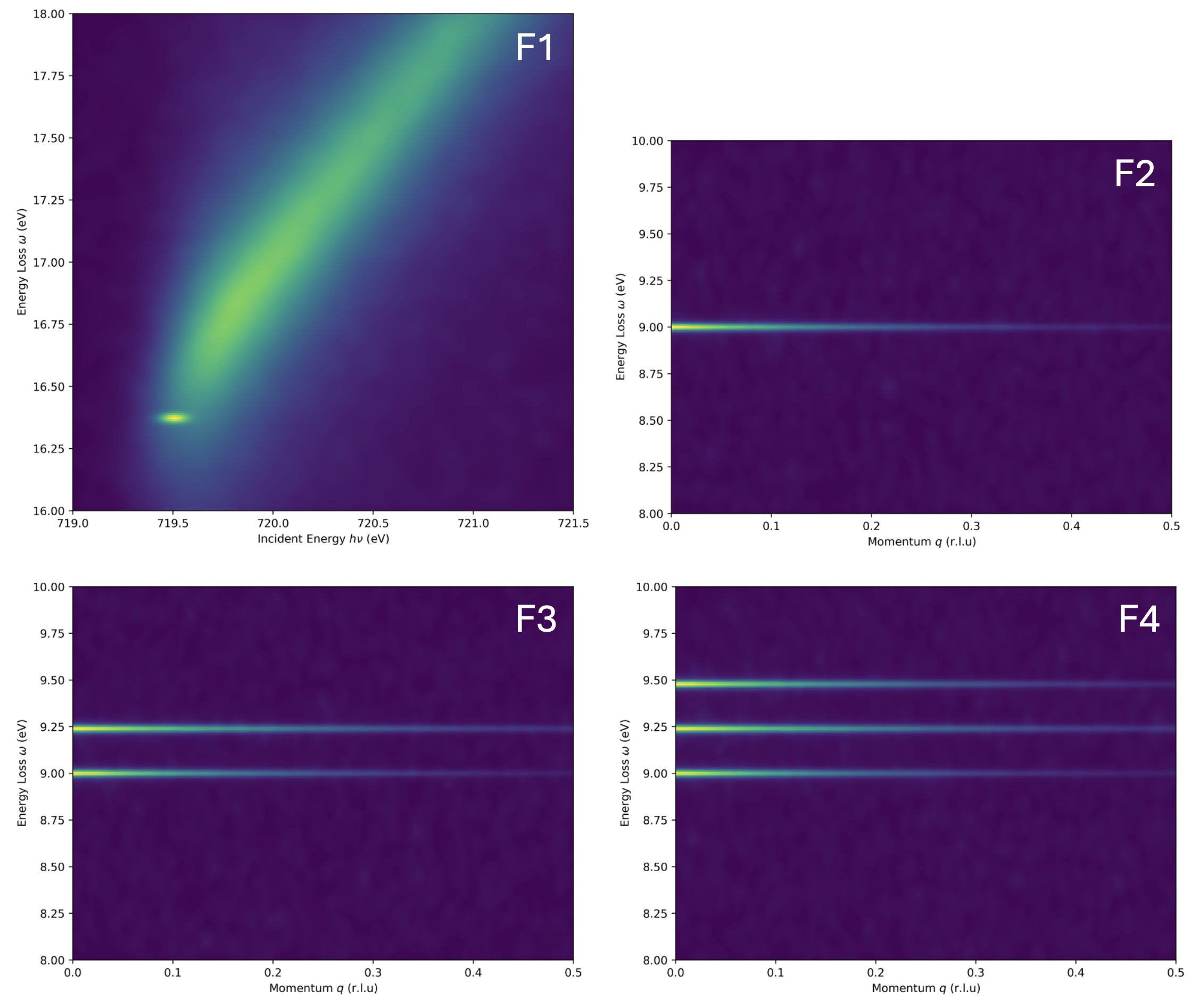}
  \caption{\textbf{Question domain F} (RIXS energy determination), including determination of feature energy (\texttt{F1}), single phonon energy (\texttt{F2}), phonon overtone energy (\texttt{F3}), and second phonon overtone energy (\texttt{F4}).}
  \label{fig:f}
\end{figure}

\begin{figure}[h!]
  \centering
  \includegraphics[width=\linewidth]{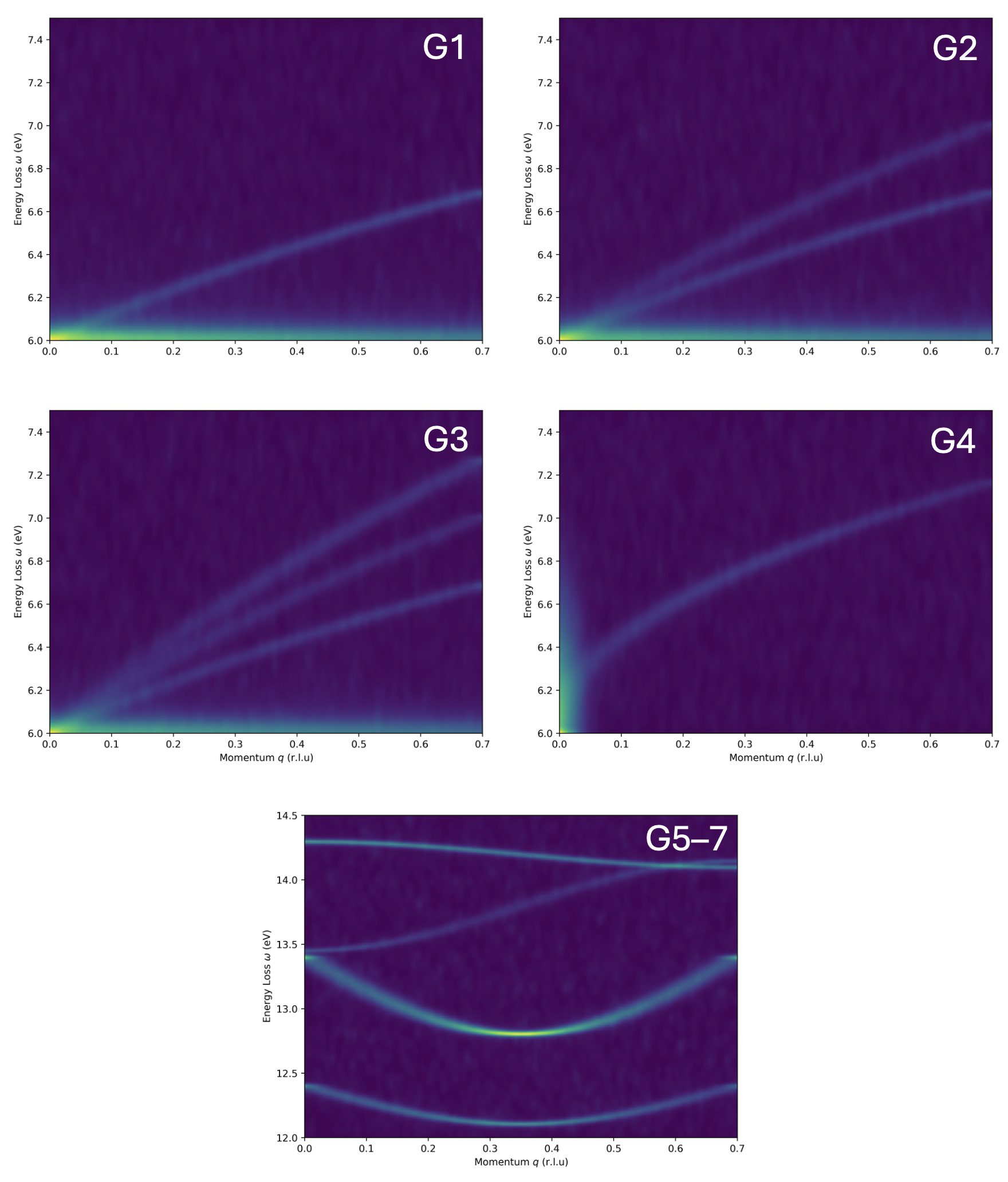}
  \caption{\textbf{Question domain G} (RIXS dispersion tracing), including tracing single plasmon dispersion (\texttt{G1}), likewise done while differentiating between two (\texttt{G2}) and three (\texttt{G3}) plasmon dispersions; single magnon dispersion (\texttt{G4}); tracing one of three orbiton dispersions (\texttt{G5–G7}).}
  \label{fig:g}
\end{figure}

\begin{figure}[h!]
  \centering
  \includegraphics[width=0.9\linewidth]{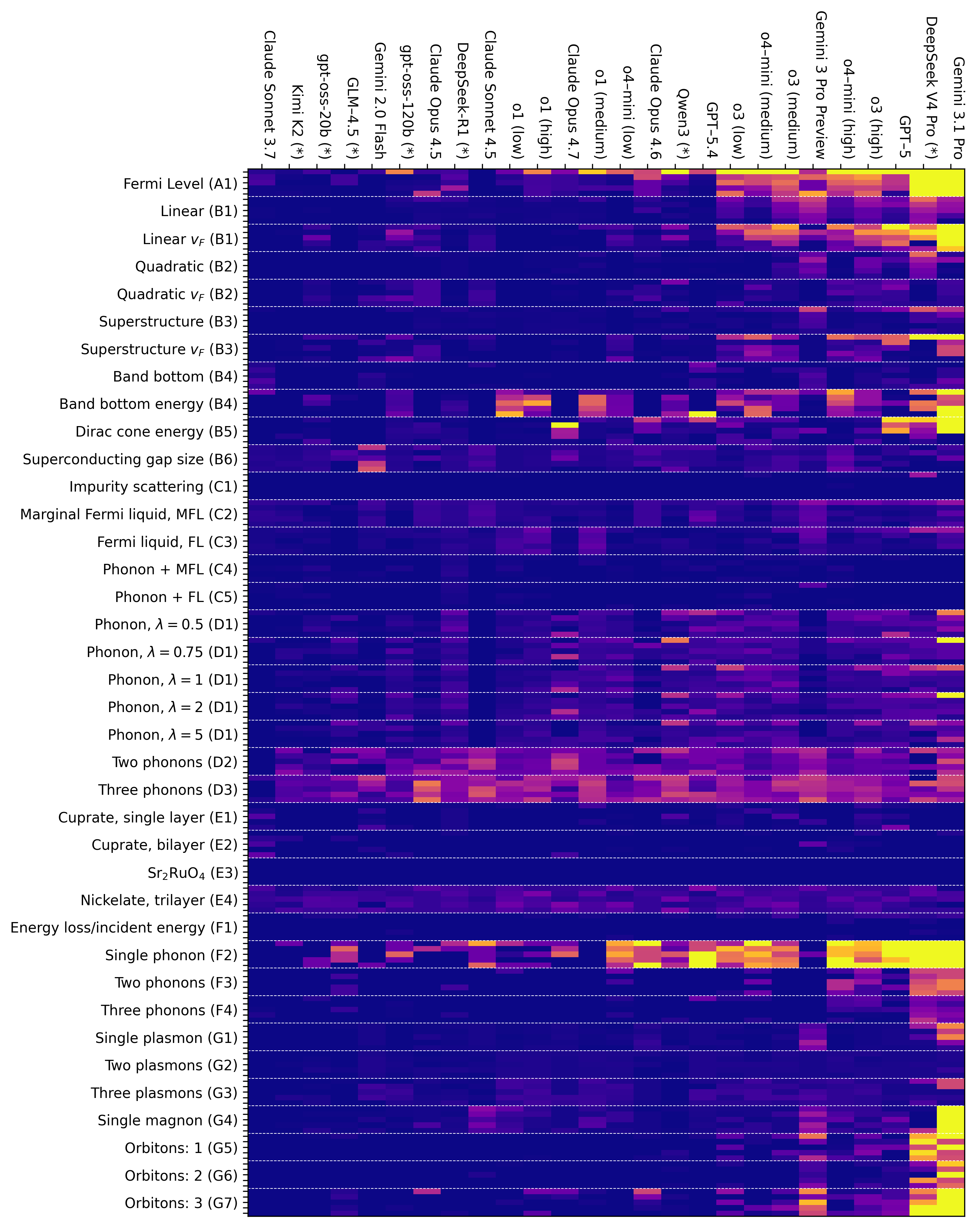}
  \caption{Scores from \textbf{direct prompting}, excluding handwritten code. Within each category (e.g. \texttt{A1}), the amount of noise (ARPES) or convolution (RIXS) increases as the plot descends. The colorplot ranges from 0 (deep purple) to 1 (bright yellow).}
  \label{fig:direct}
\end{figure}

\begin{figure}[h!]
  \centering
  \includegraphics[width=\linewidth]{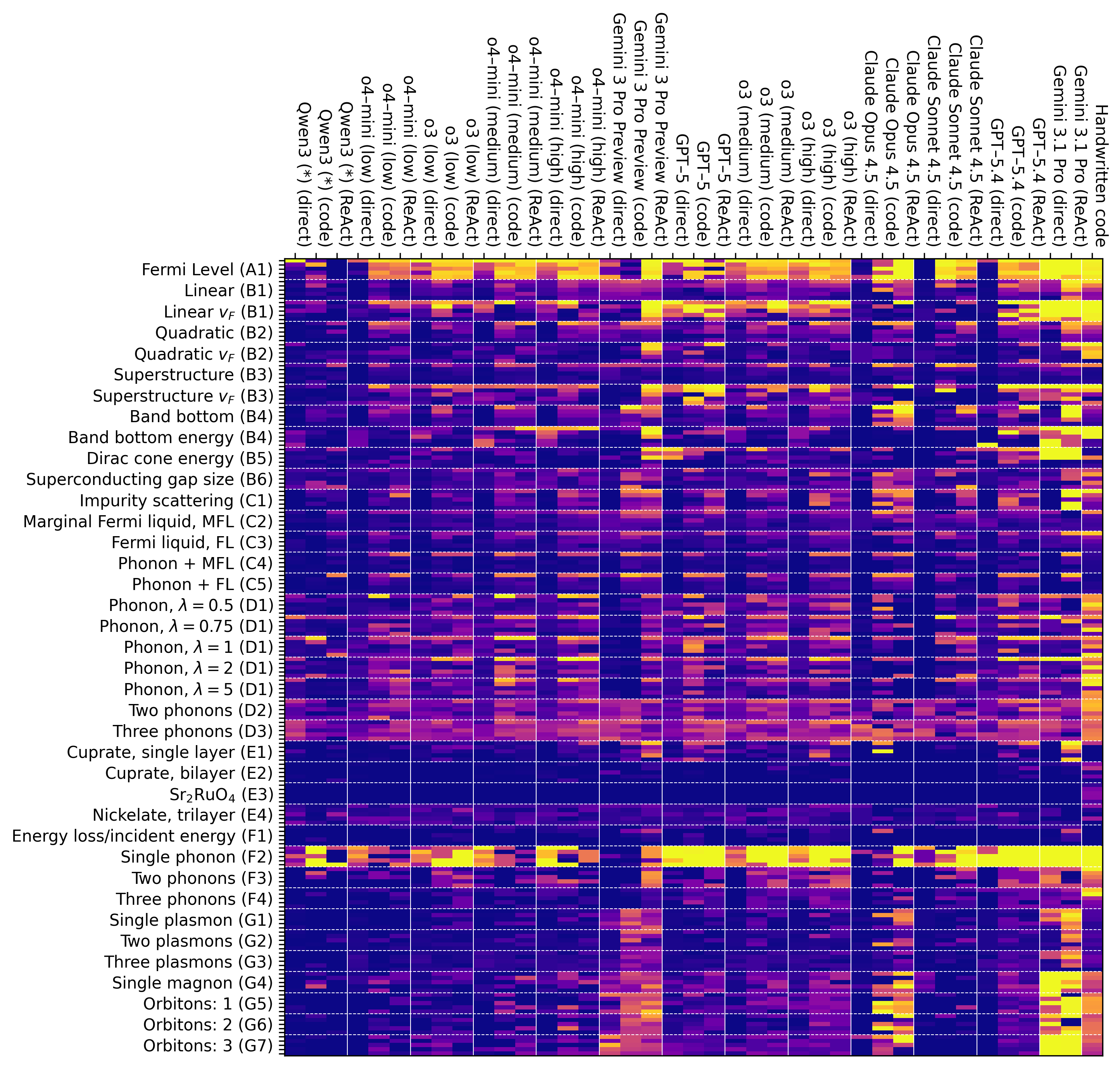}
  \caption{Scores from \textbf{direct and advanced prompting}. Within each question domain (e.g. \texttt{A1}), the amount of noise (ARPES) or convolution (RIXS) increases as the plot descends. The colorplot ranges from 0 (deep purple) to 1 (bright yellow).}
  \label{fig:advanced}
\end{figure}

Sample, handwritten code used as an equivalent non-agentic comparison is located towards the end of \texttt{client/init.py} in the repository, and is summarized here. The handwritten code draws from a toolbox of common algorithmic fitting methods, often written in laboratory–specific software packages and macros, and translated into python script by the graduate student. In cases where multiple methods are available, such as alternate curve fit profiles, we ran code for all options and chose the best score for each question. In particular, any curve fit with a Lorentzian has also been fit with Voigt profiles (Lorentzians convolved with a Gaussian to reflect instrument convolution), and Doniach–Sunjic line shapes (asymmetric Lorentzians) \citep{doniach1970}. \begin{itemize}
    \item \texttt{A1}. Intensity was summed across all momenta to yield curves that depend only on energy (these are known as energy distribution curves, or EDCs). The EDC was then fit with a Fermi–Dirac function (see caption in Fig.~\ref{fig:a}) to retrieve the Fermi level, $\mu$.
    \item \texttt{B1-B3} and \texttt{B5}. Spectra were sliced at each individual energy level to yield momentum distribution curves (MDCs, momentum-dependent counterparts to EDCs). Each MDC was fit with Lorentzian(s), and the process repeated across energies to extract their dispersions $\epsilon(k)$. Similar EDC analysis was used for \texttt{B4} and \texttt{B6}. In \texttt{B6}, which addressed EDCs in the superconducting state, line shapes from \citep{norman1998} were also used.
    \item \texttt{C1-C5}. The same MDC analysis of the \texttt{B} series of questions were used to extract line widths $\Sigma''(\omega)$, which are related to the half–width half–maxima (HWHM) $\gamma$ of Lorentzians. The main source of error in this analysis is that momentum and energy convolutions artificially increase HWHM; retrieval of actual HWHM through deconvolution or other procedures is an active area of research in the field.
    \item \texttt{D1-D3}. Line width analysis of the \texttt{C} series of questions was first carried out. Phonon frequencies were heuristically defined as those at which line width increased most rapidly with energy, within certain energy bounds. Alternatively, multiple line fits were used on the dispersion, and phonon frequencies defined at the energies where the (straight) lines intersected.
    \item \texttt{E1-E4}. Fermi surface maps were sliced at fixed values of $k_x$ and the resulting intensities were fit with Lorentzians whose peaks trace the locus of the Fermi surface, whose area is related to the doping level (see caption in Fig.~\ref{fig:e}). Trapezoidal integration and parametric fits are used.
    \item \texttt{F1-4}. Intensity was summed across all momenta to yield curves that depend only on energy (these are known as energy distribution curves, or EDCs). The EDC was then fit with a Lorentzian function (or multiple Lorentzians) to retrieve the relevant peak(s).
    \item \texttt{G1-7}. Similar to \texttt{B4}, with multiple peak fitting where required.
\end{itemize}

Note that the standard of the field is significantly higher than the given code, but the present standard of code was chosen for two reasons: (1) professional data analysis in ARPES and RIXS requires almost continual human intervention (e.g. judging goodness of fit, removing erroneous data), whereas we prefer that the code be self–contained and transparent, and (2) what constitutes the standard of the field also depends on the specific analysis task at hand; being an inverse problem with noise, it remains an active area of research across many diverse fields.

\subsection*{Reproducibility} \label{rep}

In order to reproduce the evaluation \textit{completely} from scratch, one should in principle query all questions to all models through the user's own private API(s) and server, a procedure explained in \texttt{readme\_verbose.md} in the code (not to be confused with the \texttt{README.md} data card). Because this is time–consuming and expensive, we have provided all model responses in \href{https://huggingface.co/datasets/jingene/CMESpectrumBench}{\textcolor{blue}{this repository}} so that relevant figures (Figs.~\ref{fig:leaderboards}, \ref{fig:tiers}, \ref{fig:ablations}) may be reproduced from previously collected responses. The user may also choose to inspect model responses using the \texttt{plot()} function in \texttt{eval.py}. \\

The repository can be hosted locally or on a cloud server (Codespaces was used for this evaluation). After installing the relevant dependencies (see \texttt{readme\_verbose.md}), the user should navigate to the folder \texttt{client/} and run \texttt{eval.py}. This calls \texttt{ask.py}, which uses \texttt{init.py} to generate the relevant spectra and prompts required for evaluation. In particular, the functions \texttt{ask\_batch()} and \texttt{ask\_batch\_varied()} in \texttt{ask.py} are run to generate the spectra, prompts, and answers required for direct query and advanced prompting respectively (for all questions), while \texttt{score\_models()} and \texttt{score\_categories\_resolution\_iter()} in \texttt{eval.py} score all models at standard resolution, and a selected model (Gemini 3.1 Pro) for varying resolutions, respectively. \\

Details of models evaluated in the leaderboard (Fig.~\ref{fig:leaderboards}) are stated here: \\

\begin{table}[h]
\centering
\begin{tabular}{|l|l|l|}
\hline
\textit{Model display name} & \textit{Model API name} & \textit{Context window (tokens)} \\ \hline
Claude Sonnet 3.7 & \texttt{claude-3-7-sonnet-20250219} & 200k \\
Claude Sonnet 4.5 & \texttt{claude-4-5-sonnet-20250219} & 200k \\
Claude Opus 4.5 & \texttt{claude-opus-4-5} & 200k \\
Claude Opus 4.6 & \texttt{claude-opus-4-6} & 1 million \\
Claude Opus 4.7 & \texttt{claude-opus-4-7} & 1 million \\
DeepSeek–R1 & \texttt{deepseek-reasoner} & 128k \\
DeepSeek V4 Pro & \texttt{deepseek-v4-pro} & 1 million \\
Gemini 2.0 Flash & \texttt{gemini-2.0-flash} & 1 million \\
Gemini 3 Pro Preview & \texttt{gemini-3-pro-preview} & 1 million \\
Gemini 3.1 Pro & \texttt{gemini-3.1-pro-preview} & 1 million \\
GLM–4.5 & \texttt{glm-4.5} & 131k \\
GPT–5 & \texttt{gpt-5} & 400k \\
GPT–5.4 & \texttt{gpt-5.4} & 1 million \\
gpt–oss–20b & \texttt{gpt-oss-20b} & 128k \\
gpt-oss-120b & \texttt{gpt-oss-120b} & 128k \\
Kimi K2 & \texttt{kimi-k2} & 256k \\
o1 & \texttt{o1} & 200k \\
o3 & \texttt{o3} & 200k \\
o4–mini & \texttt{o4-mini} & 200k \\
Qwen3 & \texttt{qwen3-235b-thinking} & 256k \\
\hline
\end{tabular}
\end{table}

\newpage
\section*{Appendix G\quad Question Domains and Full Scores} \label{appendixG}

The remaining pages of this text are a catalog of spectra with minimum noise and convolution, representative of question domains, and breakdowns of scores for all models.

%%%%%%%%%%%%%%%%%%%%%%%%%%%%%%%%%%%%%%%%%%%%%%%%%%%%%%%%%%%%

\clearpage
\newpage
\section*{NeurIPS Paper Checklist}

\begin{enumerate}

\item {\bf Claims}
    \item[] Question: Do the main claims made in the abstract and introduction accurately reflect the paper's contributions and scope?
    \item[] Answer: \answerYes{} % Replace by \answerYes{}, \answerNo{}, or \answerNA{}.
    \item[] Justification: We make three claims in the abstract: (1) novelty/properties of our benchmark, (2) that existing LLMs can perform comparably to graduate students when problems are rephrased as code generation problems, and (3) because future foundation models in physics must ingest and digest data directly (rather than interfacing with them through code), models' lower performance using direct query, and more concerningly under increased resolution (context length) are problems that we have identified as further avenues of research. These claims are fortified by the narrative of our paper.

\item {\bf Limitations}
    \item[] Question: Does the paper discuss the limitations of the work performed by the authors?
    \item[] Answer: \answerYes{} % Replace by \answerYes{}, \answerNo{}, or \answerNA{}.
    \item[] Justification: We discuss the limitations of our approach in the final subsection of our paper.

\item {\bf Theory assumptions and proofs}
    \item[] Question: For each theoretical result, does the paper provide the full set of assumptions and a complete (and correct) proof?
    \item[] Answer: \answerNA{} % Replace by \answerYes{}, \answerNo{}, or \answerNA{}.
    \item[] Justification: This paper does not contain theoretical results.

    \item {\bf Experimental result reproducibility}
    \item[] Question: Does the paper fully disclose all the information needed to reproduce the main experimental results of the paper to the extent that it affects the main claims and/or conclusions of the paper (regardless of whether the code and data are provided or not)?
    \item[] Answer: \answerYes{} % Replace by \answerYes{}, \answerNo{}, or \answerNA{}.
    \item[] Justification: A subsection titled `Reproducibility' in Appendix F addresses how results may be reproduced from existing collected model responses.

\item {\bf Open access to data and code}
    \item[] Question: Does the paper provide open access to the data and code, with sufficient instructions to faithfully reproduce the main experimental results, as described in supplemental material?
    \item[] Answer: \answerYes{} % Replace by \answerYes{}, \answerNo{}, or \answerNA{}.
    \item[] Justification: Code is provided in supplementary material, and data is hosted on Huggingface.

\item {\bf Experimental setting/details}
    \item[] Question: Does the paper specify all the training and test details (e.g., data splits, hyperparameters, how they were chosen, type of optimizer) necessary to understand the results?
    \item[] Answer: \answerYes{} % Replace by \answerYes{}, \answerNo{}, or \answerNA{}.
    \item[] Justification: Experimental settings are included in the code, and the exact number of responses sampled per model and question can be viewed by inspecting the dataset.

\item {\bf Experiment statistical significance}
    \item[] Question: Does the paper report error bars suitably and correctly defined or other appropriate information about the statistical significance of the experiments?
    \item[] Answer: \answerYes{} % Replace by \answerYes{}, \answerNo{}, or \answerNA{}.
    \item[] Justification: Error bars are provided for ablation tests on resolution but not for the main leaderboard, where error propagation is not well–defined, and sampling every question enough times to determine the distribution of answers and its corresponding standard deviation (i.e. whether the distribution of scores is Gaussian) is too costly and time–consuming.

\item {\bf Experiments compute resources}
    \item[] Question: For each experiment, does the paper provide sufficient information on the computer resources (type of compute workers, memory, time of execution) needed to reproduce the experiments?
    \item[] Answer: \answerNo{} % Replace by \answerYes{}, \answerNo{}, or \answerNA{}.
    \item[] Justification: As models are queried by APIs, experimental compute resources are not known, and may vary widely depending on individual rate limits.
    
\item {\bf Code of ethics}
    \item[] Question: Does the research conducted in the paper conform, in every respect, with the NeurIPS Code of Ethics \url{https://neurips.cc/public/EthicsGuidelines}?
    \item[] Answer: \answerYes{} % Replace by \answerYes{}, \answerNo{}, or \answerNA{}.
    \item[] Justification: We have reviewed the code of ethics and ensured compliance with it.

\item {\bf Broader impacts}
    \item[] Question: Does the paper discuss both potential positive societal impacts and negative societal impacts of the work performed?
    \item[] Answer: \answerNA{} % Replace by \answerYes{}, \answerNo{}, or \answerNA{}.
    \item[] Justification: As our work concerns scientific tools that measure abstract quantities in condensed matter physics very much removed from everyday properties, we do not believe there is a discernible societal impact.
    
\item {\bf Safeguards}
    \item[] Question: Does the paper describe safeguards that have been put in place for responsible release of data or models that have a high risk for misuse (e.g., pre-trained language models, image generators, or scraped datasets)?
    \item[] Answer: \answerNA{} % Replace by \answerYes{}, \answerNo{}, or \answerNA{}.
    \item[] Justification: For the same reason as the previous question, we do not believe the work poses such risks.

\item {\bf Licenses for existing assets}
    \item[] Question: Are the creators or original owners of assets (e.g., code, data, models), used in the paper, properly credited and are the license and terms of use explicitly mentioned and properly respected?
    \item[] Answer: \answerYes{} % Replace by \answerYes{}, \answerNo{}, or \answerNA{}.
    \item[] Justification: The code provided was written by the authors of this work, which generated synthetic data based on abstract scientific formulae, while LLM models were queried using APIs and paid for by the authors. The code licence is CC BY 4.0 and is available at \href{https://huggingface.co/datasets/jingene/CMESpectrumBench}{\textcolor{blue}{\texttt{https://huggingface.co/datasets/jingene/CMESpectrumBench}}}.

\item {\bf New assets}
    \item[] Question: Are new assets introduced in the paper well documented and is the documentation provided alongside the assets?
    \item[] Answer: \answerYes{} % Replace by \answerYes{}, \answerNo{}, or \answerNA{}.
    \item[] Justification: The dataset, which was generated by the authors, is uploaded on Hugging Face (see above question).

\item {\bf Crowdsourcing and research with human subjects}
    \item[] Question: For crowdsourcing experiments and research with human subjects, does the paper include the full text of instructions given to participants and screenshots, if applicable, as well as details about compensation (if any)? 
    \item[] Answer: \answerNA{} % Replace by \answerYes{}, \answerNo{}, or \answerNA{}.
    \item[] Justification: No crowdsourcing or research with human subjects was involved. Note that `handwritten code' (mentioned, for example, in the leaderboard) was written by an author of this publication.

\item {\bf Institutional review board (IRB) approvals or equivalent for research with human subjects}
    \item[] Question: Does the paper describe potential risks incurred by study participants, whether such risks were disclosed to the subjects, and whether Institutional Review Board (IRB) approvals (or an equivalent approval/review based on the requirements of your country or institution) were obtained?
    \item[] Answer: \answerNA{} % Replace by \answerYes{}, \answerNo{}, or \answerNA{}.
    \item[] Justification: No crowdsourcing or research with human subjects was involved.

\item {\bf Declaration of LLM usage}
    \item[] Question: Does the paper describe the usage of LLMs if it is an important, original, or non-standard component of the core methods in this research? Note that if the LLM is used only for writing, editing, or formatting purposes and does \emph{not} impact the core methodology, scientific rigor, or originality of the research, declaration is not required.
    %this research? 
    \item[] Answer: \answerYes{} % Replace by \answerYes{}, \answerNo{}, or \answerNA{}.
    \item[] Justification: The usage of LLMs in this paper is exclusively as a test subject: we are measuring their performance on our benchmark.

\end{enumerate}

\end{document}